\documentclass[sigconf,balance=false]{acmart}
\renewcommand\footnotetextcopyrightpermission[1]{} 

\AtBeginDocument{%
  \providecommand\BibTeX{{%
    \normalfont B\kern-0.5em{\scshape i\kern-0.25em b}\kern-0.8em\TeX}}}

\setcopyright{acmcopyright}
\copyrightyear{2018}
\acmYear{2018}
\acmDOI{XXXXXXX.XXXXXXX}

\acmConference[CCS '24]{The 31th ACM Conference on Computer and Communications Security}{October 14-18, 2024}{Salt Lake City, U.S.A.}
\usepackage{algorithm,algpseudocode}
\usepackage{tikz}
\usepackage{bbm}
\usepackage{multirow}
\usepackage{fontawesome5,pifont} 
\usepackage{subcaption}
\usepackage{enumitem}

\newtheorem{Thm}{Theorem}[section]
\newtheorem{Lem}[Thm]{Lemma}

\newtheorem{Exm}[Thm]{Example}

\newcounter{protocol}
\makeatletter
\newenvironment{protocol}[1][htb]{%
    \let\c@algorithm\c@protocol
    \renewcommand{\ALG@name}{Protocol}
    \begin{algorithm}[#1]%
    }{\end{algorithm}
}
\makeatother

\newcounter{protocol*}
\makeatletter
\newenvironment{protocol*}[1][htb]{%
    \let\c@algorithm\c@protocol
    \renewcommand{\ALG@name}{Protocol}
    \begin{algorithm*}[#1]%
    }{\end{algorithm*}
}
\makeatother

\newcommand{\ceil}[1]{\left\lceil#1 \right\rceil}

\newcommand{\round}[1]{\left\lfloor#1 \right\rceil}
\newcommand{\abs}[1]{\left|#1 \right|}

\newcommand{\innerproduct}[2]{\left\langle#1, #2\right\rangle}

\newcommand{\me}[2]{\widetilde{#1}\left(#2\right)}

\renewcommand{\bf}[1]{\ifmmode\mathbf{#1}\else\textbf{#1}\fi}

\renewcommand{\tt}[1]{\ensuremath{\texttt{#1}}}
\newcommand{\com}[1]{\ensuremath{\dbbracket{#1}}}

\newcommand{\negl}[1]{\ensuremath{\tt{negl}\left({#1}\right)}}

\newcommand{\bracket}[1]{\ensuremath{\left[#1\right]}}
\newcommand{\dbbracket}[1]{\ensuremath{\left\llbracket#1\right\rrbracket}}
\newcommand{\cbracket}[1]{\ensuremath{\left\{#1\right\}}}

\def\F{\mathbb{F}}
\def\G{\mathbb{G}}

\def\1{\mathbbm{1}}

\newcommand{\case}[2][lllllllllllllllllllllllllllllllllllll]{\left\{\begin{array}{#1}#2 \\ \end{array}\right.}

\usepackage{mdframed}
\definecolor{lightgray}{gray}{0.9}

\usepackage{pifont}

\begin{document}
\mdfsetup{linewidth=1pt}

\title{zkLLM: Zero Knowledge Proofs for Large Language Models}

\author{Haochen Sun}
\email{haochen.sun@uwaterloo.ca}
\affiliation{%
  \institution{University of Waterloo}
  \streetaddress{200 University Ave W}
  \city{Waterloo}
  \state{Ontario}
  \country{Canada}
  \postcode{N2L 3G1}}

\author{Jason Li}
\email{j2643li@uwaterloo.ca}
\affiliation{%
  \institution{University of Waterloo}
  \streetaddress{200 University Ave W}
  \city{Waterloo}
  \state{Ontario}
  \country{Canada}
  \postcode{N2L 3G1}}

\author{Hongyang Zhang}
\email{hongyang.zhang@uwaterloo.ca}
\affiliation{%
  \institution{University of Waterloo}
  \streetaddress{200 University Ave W}
  \city{Waterloo}
  \state{Ontario}
  \country{Canada}
  \postcode{N2L 3G1}}

\begin{abstract}
    The recent surge in artificial intelligence (AI), characterized by the prominence of large language models (LLMs), has ushered in fundamental transformations across the globe. However, alongside these advancements, concerns surrounding the legitimacy of LLMs have grown, posing legal challenges to their extensive applications. Compounding these concerns, the parameters of LLMs are often treated as intellectual property, restricting direct investigations.

    In this study, we address a fundamental challenge within the realm of AI legislation: the need to establish the authenticity of outputs generated by LLMs. To tackle this issue, we present \textbf{zkLLM}, which stands as the inaugural specialized zero-knowledge proof tailored for LLMs to the best of our knowledge. Addressing the persistent challenge of non-arithmetic operations in deep learning, we introduce \texttt{tlookup}, a parallelized lookup argument designed for non-arithmetic tensor operations in deep learning, offering a solution with no asymptotic overhead. Furthermore, leveraging the foundation of \texttt{tlookup}, we introduce \texttt{zkAttn}, a specialized zero-knowledge proof crafted for the attention mechanism, carefully balancing considerations of running time, memory usage, and accuracy.

    Empowered by our fully parallelized CUDA implementation, zkLLM emerges as a significant stride towards achieving efficient zero-knowledge verifiable computations over LLMs. Remarkably, for LLMs boasting 13 billion parameters, our approach enables the generation of a correctness proof for the entire inference process in under 15 minutes. The resulting proof, compactly sized at less than 200 kB, is designed to uphold the privacy of the model parameters, ensuring no inadvertent information leakage.
\end{abstract}

\maketitle
\section{Introduction} \label{sec:introduction}


The recent surge in artificial intelligence (AI), particularly with the advent of Large Language Models (LLMs) \cite{gpt3, gpt4, llama, llama2, palm, gemini}, has profoundly transformed the world. However, these technological advances have also raised concerns about the legitimacy of these groundbreaking models, challenging the legal underpinnings of their extensive applications. For instance, in December 2023, the New York Times filed a lawsuit against OpenAI and Microsoft, accusing them of using copyrighted material from the newspaper to train their chatbots. In October 2023, President Biden issued an executive order to address both the "myriad benefits" and "substantial risks" posed by AI. As laws and regulations around LLMs evolve and tighten, developing practical tools to verify the legitimacy of these models has become crucial.

Consider the auditing process of a newly-released LLM, which is hosted on a cloud service (e.g., Microsoft Azure) with API access. Law enforcement queries the model using designated prompts to test if the LLM generates illegal output (e.g., untrue, violence-prompting, or racist). In the stringent legal context, the authenticity of the output must be established to exclude the possibility of cheating by manipulating the generated texts. On the other hand, although the architectures are typically described in technical reports, the trained parameters are concealed as the AI developers' intellectual properties, making direct examination of the model parameters impossible. This dilemma calls for the application of zero-knowledge proofs (ZKPs), which allow for verifiable computations over the neural networks while disclosing no information about the neural network parameters \cite{zkcnn, zen, vcnn, pvcnn, zkml, mystique, ezdps}.


However, adapting existing ZKP techniques to modern LLMs, characterized by their immense scale, presents significant challenges. These models require substantial computational resources, which general-purpose ZKP frameworks \cite{pinocchio,groth16,plonk,halo,halo-inf,DBLP:journals/joc/BitanskyCIOP22,hyperplonk,ligero,ligero++}, often unaware of LLM structure and limited in parallel computation support, struggle to provide.

While early research has explored specialized cryptographic protocols for specific neural network architectures like convolutional neural networks (CNNs) \cite{vcnn, pvcnn, zkcnn}, LLMs' complex internal structures necessitate further innovation in ZKP protocol design. This innovation is vital to avoid the excessive overhead typical of general-purpose ZKPs. LLMs involve many non-arithmetic operations, such as GELU \cite{gelu} and SwiGLU \cite{swiglu} activation functions, which only partially align with current ZKP methods. Lookup arguments, trading memory consumption for faster runtimes, have been introduced \cite{zkml} to handle these nonlinearities, but their straightforward application raises questions about manageable memory overhead. 

Moreover, the attention mechanism in LLMs \cite{attention}, which is inherently multivariate and often employs the Softmax function, requires a tailored ZKP protocol design for effective management of proof overhead. Tackling this mechanism within ZKPs is challenging, particularly as its components are not typically found in previously explored neural network architectures, such as MLPs and CNNs. In these traditional models, Softmax functions are usually placed after the output layer and are therefore not considered in prior works on zero-knowledge verifiable deep learning. This setup is in stark contrast with LLMs, where Softmax functions are used extensively across multiple layers. This prevalent use in LLMs necessitates a more refined approach in ZKP design to ensure both precise and efficient zero-knowledge verification, especially given the unique challenges presented by the attention mechanism.

In response to these challenges, we present \textbf{zkLLM}, the inaugural ZKP scheme specifically designed for LLMs. zkLLM empowers LLM owners to validate the integrity of inference outcomes to stakeholders, such as law enforcement agencies, thereby streamlining investigations involving LLMs while safeguarding intellectual property. Our key contributions are:

\begin{itemize}
    \item We propose \texttt{tlookup}, a unique ZKP protocol for universal non-arithmetic operations in deep learning, to tackle the persistent challenge of verifying such operations (e.g., activation functions). \texttt{tlookup} adeptly handles overhead in two ways: analytically, it adds no asymptotic overhead in memory complexity or running time; practically, its design promotes a high level of parallelization, fully leveraging parallel computing resources (like GPUs) commonly used in LLM computing environments.
    \item We introduce \texttt{zkAttn}, a ZKP specifically crafted for attention mechanisms in LLMs. Building upon \texttt{tlookup} and enhancing its capabilities, \texttt{zkAttn} mitigates the accuracy degradation and high overheads linked with bit-decompositions and polynomial approximations. It also removes the necessity to list all multivariate input-output pairs, a prerequisite in lookup-based methods, by harnessing the mathematical properties of the attention mechanism. This strategy strikes a balance between running time, memory usage, and accuracy, while maintaining security and privacy standards.
    \item Our efficient CUDA implementation, in conjunction with the aforementioned technical advancements, positions zkLLM as the trailblazing ZKP for LLMs of sizes up to 13 billion parameters. zkLLM achieves reasonable proving times of 1-15 minutes and produces compact proofs smaller than 200kB. These proofs can be verified within 1-3 seconds by the verifier and guarantee no exposure of model parameters.
\end{itemize}

\section{Technical overview} \label{sec:overview}

Compared with general-purpose counterparts, an efficient zero-knowledge proof system specialized for deep learning hinges critically upon two key requirements:
\begin{itemize}
    \item The capability for extensive parallelization (for example, using CUDA), which allows for the handling of proofs for the entire computational process in a reasonable timeframe.
    \item The adept handling of non-arithmetic operations, encompassing activation functions among others.
\end{itemize}

Although sumcheck-based protocols are known to be compatible with tensor structures common in deep learning computations \cite{safetynets,zkcnn} , traditionally, they have depended on bit-decomposition methods for non-arithmetic operations. This dependence leads to an increase in prover overhead and restricts the variety of non-arithmetic operations that can be supported. In response, we have developed a novel sumcheck-based protocol for lookup arguments over tensors.

Our design capitalizes on the following fact: for $\mathbf{S}\in \F^D$ and $\mathbf{T} \in \F^N$, the set inclusion $\mathbf{S}\subseteq\mathbf{T}$ holds if and only if there is an $\mathbf{m}$ such that $\sum_{i\in [D]} (X + \mathbf{S}_i)^{-1} \equiv \sum_{i\in [N]}\mathbf{m}_i (X+\mathbf{T}_i)^{-1}$ as rational functions over $X \in \F$ \cite{DBLP:journals/iacr/Habock22a}. This equivalence can be verified by evaluating both expressions at a single point $X \gets \beta$, randomly selected by the verifier. Furthermore, $\mathbf{m}$ can be computed in $O(D)$ time using straightforward counting. Hence, by calculating the elementwise multiplicative inversions $\mathbf{A} \gets (\beta + \mathbf{S})^{-1}$ and $\mathbf{B} \gets (\beta + \mathbf{T})^{-1}$, we can parallelize the sumcheck protocol for the identity 
\begin{equation}
    (\mathbf{A}.\text{sum()} = \mathbf{m}^\top\mathbf{B}) \land (\mathbf{A} \odot (\beta + \mathbf{S}) = \mathbf{1}) \land (\mathbf{B} \odot (\beta + \mathbf{T}) = \mathbf{1})
\end{equation}
This approach stands in contrast to the sequential lookup arguments \cite{plookup, caulk, caulk+, flookup, baloo, cq} that are based on univariate polynomials.

\begin{figure}
    \centering
    \includegraphics[page=1,width=\linewidth]{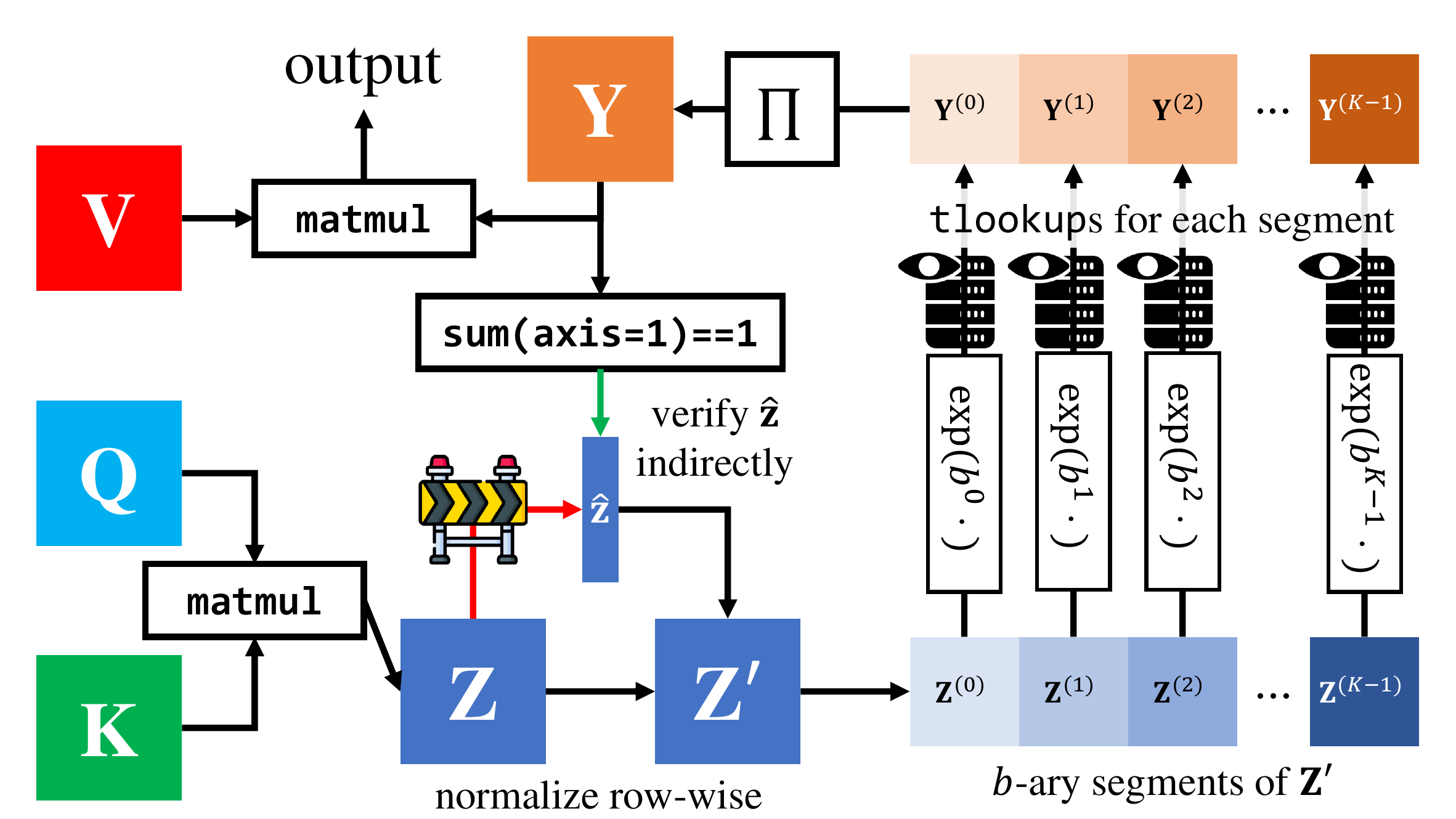}
    \caption{Overview of \tt{zkAttn} for \eqref{eq:attn}. }
    \label{fig:zkattn-overview}
\end{figure}

For the attention mechanism widely applied in modern LLMs, represented by the equation
\begin{equation}
    \text{Attention}(\mathbf{Q}, \mathbf{K}, \mathbf{V}) := \text{Softmax}\left(\frac{\mathbf{Q}\mathbf{K}^\top}{\sqrt{d}}\right)\mathbf{V},\label{eq:attn}
\end{equation}
the direct application of lookup arguments, such as \texttt{tlookup}, presents the impractical challenge of compiling all input-output pairs into a lookup table due to the multivariate nature of the attention mechanism. To achieve zero-knowledge verifiability with limited overhead for the attention mechanism, we introduce \texttt{zkAttn}, as depicted in Figure \ref{fig:zkattn-overview}:

\begin{enumerate}
    \item Implement the matrix multiplication between the query \(\mathbf{Q}\) and keys \(\mathbf{K}^\top\), resulting in \(\mathbf{Z}\gets \mathbf{Q}\mathbf{K}^\top\). This process is verifiable through the dedicated sumcheck protocol designed specifically for matrix multiplications.
    \item Exploit the shift-invariance property of Softmax to adjust each row of \(\mathbf{Z}\) by a constant, represented as a vector \(\hat{\mathbf{z}}\), so that \(\exp(\mathbf{Z} - \hat{\mathbf{z}}\mathbf{1}^\top)\) sums to 1 row-wise. This transformation renders the Softmax output equivalent to applying \(\exp(\cdot)\) element-wise to \(\mathbf{Z}':= \mathbf{Z} - \hat{\mathbf{z}}\mathbf{1}^\top\). However, computing \(\hat{\mathbf{z}}\) from \(\mathbf{Z}\) is highly intricate and not directly verifiable.
    \item Transform \(\mathbf{Z}'\) into negative \(K\)-digit base-\(b\) numbers, with each \(\mathbf{Z}' = -\sum_{k=0}^{K-1}b^k\mathbf{Z}^{(k)}\). By utilizing the homomorphism of \(\exp\left(\cdot\right)\), the Softmax output \(\mathbf{Y}\) is then expressed as
    \begin{equation}
        \mathbf{Y} = \exp\left(-\sum_{k=0}^{K-1}b^k\mathbf{Z}^{(k)}\right) = \prod_{k=0}^{K-1} \exp\left(-b^k\mathbf{Z}^{(k)}\right),
    \end{equation}
    with a \texttt{tlookup} installed for each term of the \(K\) in the product to handle the non-arithmetic operation.
    \item Rather than verifying the correctness of \(\hat{\mathbf{z}}\) directly, which is highly non-arithmetic, an additional check is introduced to ensure the rowwise sums of \(\mathbf{Y}\) equal 1.
    \item Implement another verifiable matrix multiplication between the Softmax output \(\mathbf{Y}\) and the values \(\mathbf{V}\).
\end{enumerate}

Note that the above overview omits details about the handling of scaling factors and quantization errors for clarity. The design of \texttt{zkAttn} manages the overhead of verifiable computation for the highly non-arithmetic operations within the attention mechanism while preserving computational accuracy.

\section{Preliminaries} \label{sec:preliminaries}

\subsection{Notations}

We represent vectors and tensors in bold font, such as $\mathbf{v}$ for vectors and $\mathbf{S}$ for tensors. Consistent with the cryptographic frameworks we utilize, we apply 0-based indexing to all mathematical structures. For simple operations and indexing over tensors, we adhere to the PyTorch conventions, using notations like $\mathbf{v}_{[i]}$ or the more concise $\mathbf{v}_i$ for elements, $\mathbf{S}_{[i, j_0:j_1, :]}$ for slicing, and $\mathbf{S}.\text{sum(axis=0)}$ for aggregation along a specified dimension. Moreover, we denote the set of non-negative integers less than a positive integer \( N \) using the compact form \( [N] \), which signifies \( \{0, 1, \dots, N-1\} \).

\subsection{Large language models, transformers and the attention mechanism}

Large Language Models (LLMs), exemplified by the GPT series \cite{gpt3, gpt4}, PaLM \cite{palm}, and LLaMa \cite{llama}, are renowned for their exceptional performance in general-purpose language understanding and generation tasks. These models are built upon the transformer architecture, which is fundamentally based on the attention mechanism \cite{attention}.

\begin{figure}
    \centering
    \begin{mdframed}[backgroundcolor=white,linewidth=1pt,roundcorner=5pt]
        \textbf{Q: }As the developer of an LLM, law enforcement seeks to verify the model’s authenticity by having me execute the inference process using their designated prompts. I aim to establish the authenticity of my output without revealing any details about the model parameters. Can ZKP be utilized in the inference process?

        \textbf{A: }Yes, ZKP can be utilized in the inference process of an LLM to establish the authenticity of the output without revealing any details about the model parameters. This allows one party to prove to another that a certain statement is true, without conveying any additional information. In this case, it can prove that the model's output came from a specific model.

        \textbf{Q: }What are some benefits of applying ZKPs in this scenario?

        \textbf{A: }Utilizing ZKPs in the inference process of an LLM can provide authenticity and privacy. It can confirm the output's origin from a specific model without revealing any model details, protecting any sensitive or proprietary information. The generated proofs are verifiable, allowing anyone to confirm the output's authenticity. Some ZKP protocols are also scalable, accommodating large models and complex computations, which is beneficial for LLMs.
    \end{mdframed}
    \caption{An example dialogue with GPT-3.5 regarding zkLLM's motivation}
    \label{fig:llm-example}
\end{figure}

\begin{figure}
    \centering
    \includegraphics[page=2,width=\linewidth]{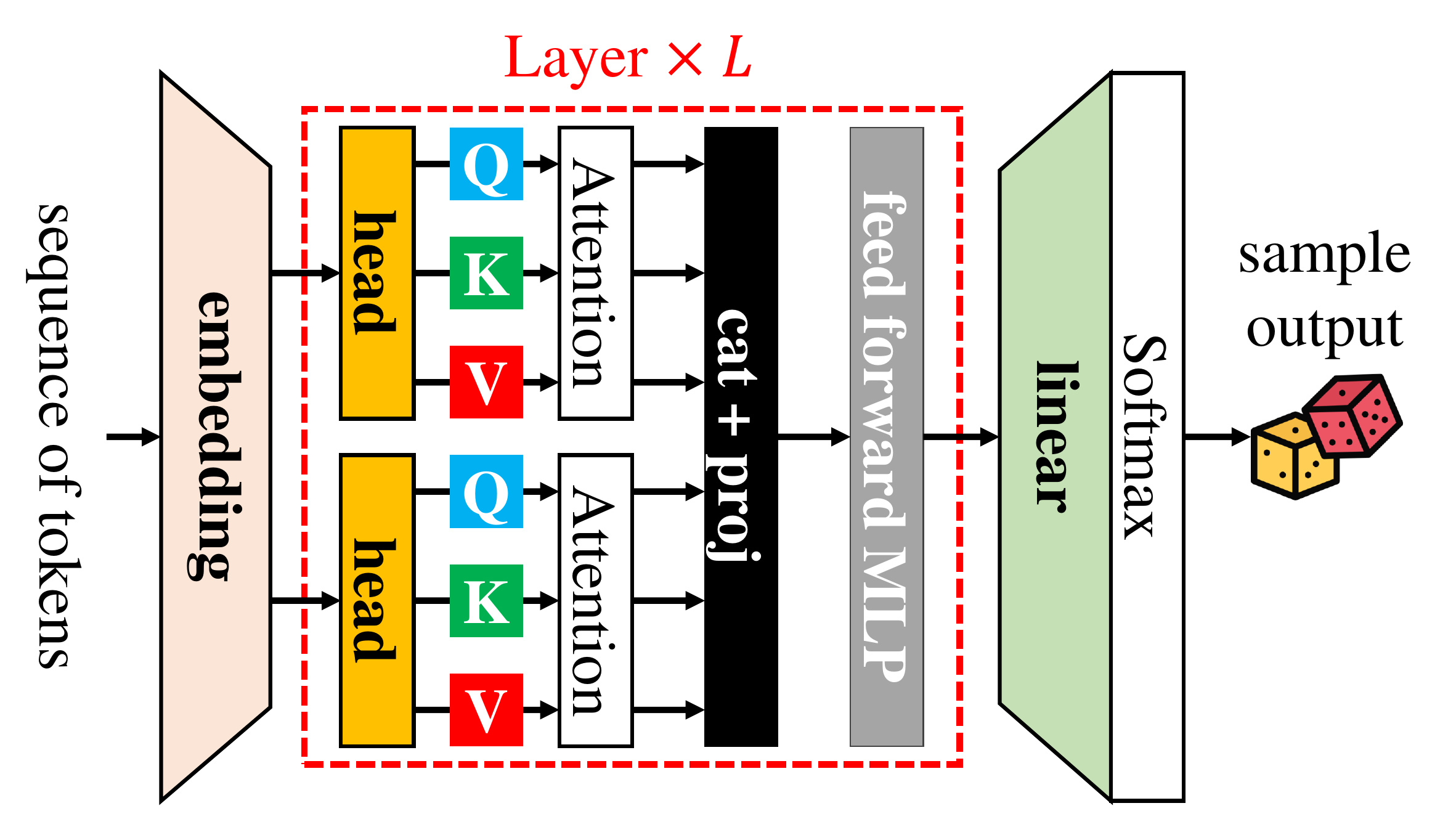}
    \caption{Typical structure of LLMs}
    \label{fig:llm-structure}
\end{figure}

As depicted in Figure \ref{fig:llm-structure}, LLMs typically consist of multiple layers that transform the embeddings of an input sequence of tokens using multi-head attention. In each attention head \(i\), parameterized by linear weights \(\mathbf{W}^Q_i\), \(\mathbf{W}^K_i\), and \(\mathbf{W}^V_i\), the queries \(\mathbf{Q}_i\gets \mathbf{X}\mathbf{W}^Q_i\), keys \(\mathbf{K}_i\gets \mathbf{X}\mathbf{W}^K_i\), and values \(\mathbf{V}_i\gets \mathbf{X}\mathbf{W}^V_i\) are computed. These components are then processed by the Attention function, concatenated, and projected using another linear weight \(\mathbf{W}^O\):
\begin{equation}
    \mathbf{O}\gets \text{Concat}_i\left(\text{Attention}\left(\mathbf{Q}_i, \mathbf{K}_i, \mathbf{V}_i\right)\right)\mathbf{W}^O,
\end{equation}
The output \(\mathbf{O}\) is subsequently processed by a feed-forward multi-layer perceptron (MLP). The activations from the final layer are then transformed into a probability distribution over the output tokens, from which the output sequence is autoregressively sampled. The Attention function, as defined in \eqref{eq:attn}, effectively mimics cognitive attention and has significantly contributed to the success of LLMs. The adaptation of the attention mechanism for zero-knowledge verifiability is a primary focus of this study.

\subsection{Sumcheck protocol, multilinear extensions and tensor operations} \label{sec:sumcheck}

The correctness of arithmetic tensor operations (e.g., matrix multiplication) is verified using the \emph{sumcheck protocol} \cite{sumcheck} over the \emph{multilinear extensions} \cite{me} of the tensors involved.

Consider a tensor $\mathbf{S}\in\F^{D_0\times D_1\times \dots \times D_{K-1}}$ discretized into a finite field $\F$ via scaling and rounding. Without loss of generality, assume that $D_k$s are all powers of 2 for $0\leq k \leq K-1$, or zero-padding may be applied. Thus, writing indices in binary format, $\mathbf{S}$ can also be considered as a function $\mathbf{S}(\cdot): {\{0, 1\}}^{\sum_{k=0}^{K-1}\log_2 D_k} \to \F$. Here, $\mathbf{S}\left(\mathbf{i}_0, \mathbf{i}_1, \dots, \mathbf{i}_{K-1}\right) = \mathbf{S}_{\left[i_0, i_1, \dots, i_{K-1}\right]}$ where $\mathbf{i}_k$ is the binary representation of any $0\leq i_k \leq D_{K-1} - 1$. A multivariate polynomial $\me{\mathbf{S}}{\cdot}: \F^{\sum_{k=0}^{K-1}\log_2 D_k} \rightarrow \F$ is a \emph{multilinear extension} of $\mathbf{S}(\cdot)$ such that $\me{\mathbf{S}}{\cdot} \equiv \mathbf{S}(\cdot)$ on $\{0, 1\}^{\sum_{k=0}^{K-1}\log_2 D_k}$, practically implemented as \begin{multline}
    \me{\mathbf{S}}{\mathbf{u}_0, \mathbf{u}_1, \dots, \mathbf{u}_{K-1}} =\\ \sum_{ \substack{\mathbf{i}_k\in{\{0, 1\}}^{\log_2 D_k}\\0\leq k \leq K-1}}\me{\mathbf{e}}{\bigoplus_{k=0}^{K-1}\mathbf{u}_k, \bigoplus_{k=0}^{K-1}\mathbf{i}_k}\mathbf{S}(\mathbf{i}_0, \mathbf{i}_1, \dots, \mathbf{i}_{K-1}), 
\end{multline} where $\me{\mathbf{e}}{\mathbf{u}, \mathbf{v}} := \sum_{i=0}^{d-1} \mathbf{u}_i\mathbf{v}_i + (1-\mathbf{u}_i)(1-\mathbf{v}_i)$ for any $d$-dimensional $\mathbf{u}, \mathbf{v}\in \F^d$, which reduces to the equality indicator $\1_{\{\mathbf{u}=\mathbf{v}\}}$ when restricted to $\mathbf{u}, \mathbf{v} \in \{0, 1\}^d$.

The correctness of a tensor operation can be expressed as equalities over the tensors. For instance, for matrix multiplication $\mathbf{C} \gets \mathbf{A}\mathbf{B}$, where $\mathbf{C}\in \F^{D_0\times D_2}$, $\mathbf{A}\in \F^{D_0\times D_1}$, and $\mathbf{B}\in \F^{D_1\times D_2}$, the correctness is characterized by $\mathbf{C}_{[i, j]} = \sum_{k=0}^{D_1-1}\mathbf{A}_{[i, k]}\mathbf{B}_{[k,j]}$ for each $i, j$, or equivalently,

\begin{equation}
    \sum_{\mathbf{k}\in \{0, 1\}^{\log_2D_1}}\left(D_1^{-1} \me{\mathbf{C}}{\mathbf{i}, \mathbf{j}} -\me{\mathbf{A}}{\mathbf{i}, \mathbf{k}}\me{\mathbf{B}}{\mathbf{k}, \mathbf{j}}\right) = 0. \label{eq:sc-intro-individual}
\end{equation}

By applying the Schwartz-Zippel Lemma \cite{DBLP:journals/jacm/Schwartz80, DBLP:conf/eurosam/Zippel79}, with high probability, \eqref{eq:sc-intro-individual} holds for all $\mathbf{i}, \mathbf{j}$ if and only if the random linear combination \begin{equation}
    \sum_{\substack{\mathbf{i}\in \{0, 1\}^{\log_2D_0}\\\mathbf{j}\in \{0, 1\}^{\log_2D_2}\\\mathbf{k}\in \{0, 1\}^{\log_2D_1}}}\me{\mathbf{e}}{\mathbf{u}_0 \oplus \mathbf{u}_2, \mathbf{i} \oplus \mathbf{j}}\underbrace{\left(D_1^{-1} \me{\mathbf{C}}{\mathbf{i}, \mathbf{j}} -\me{\mathbf{A}}{\mathbf{i}, \mathbf{k}}\me{\mathbf{B}}{\mathbf{k}, \mathbf{j}}\right)}_{\text{varies for other tensor operations}} = 0, \label{eq:sc-intro-aggr}
\end{equation} where $\me{\mathbf{e}}{\cdot}$. Thus, the prover and the verifier can execute the sumcheck protocol \cite{sumcheck}, which proves the statements in the form of \begin{equation}\label{eq:sumcheck}
    \sum_{\mathbf{i}\in \{0, 1\}^d} f(\mathbf{i}) = 0
\end{equation} for any $d$-variate polynomial ($d= \log_2D_0 + \log_2D_1 + \log_2 D_2$ in the case of \eqref{eq:sc-intro-aggr}). The prover time, proof size, and verifier time are $O(2^d)$, $O(d)$, and $O(d)$, respectively. At the end of the protocol, a claim about the value of $f(\mathbf{v})$ is made by the prover (where $\mathbf{v} \sim \F^d$ due to the randomness over the protocol execution), which is further reduced to the claimed evaluations of the multilinear extensions (i.e., $\me{\mathbf{C}}{\mathbf{v}_0, \mathbf{v}_2}$, $\me{\mathbf{A}}{\mathbf{v}_0, \mathbf{v}_1}$, $\me{\mathbf{B}}{\mathbf{v}_1, \mathbf{v}_2}$ in \eqref{eq:sc-intro-aggr} with the indices decomposed as $\mathbf{v} = \mathbf{v}_0\oplus \mathbf{v}_1 \oplus \mathbf{v}_2$ with the corresponding dimensionalities.) These claims are further verified via the proof of evaluations on the commitments of the tensors introduced in Section \ref{sec:commitment}. 

Optimized adaptations of the sumcheck protocol are designed to align with standard operations in deep learning, such as matrix multiplication \cite{DBLP:conf/crypto/Thaler13, safetynets} (see Section \ref{sec:taxonomy-matmul}) and convolution \cite{zkcnn}. The preservation of the tensor structure enables the parallelization of the proof. Moreover, zero-knowledge adaptations of the sumcheck protocol \cite{DBLP:journals/eccc/ChiesaFS17, libra, orion} have been developed to prevent any disclosure of information related to the tensors, while adding a negligible additional computational burden.

\subsection{Polynomial Commitment} \label{sec:commitment}

The binding and hiding requirements for the tensors, in the form of multilinear extensions, which are considered the intellectual properties of the prover, are achieved using polynomial commitment schemes. Specifically, the following establishes the correctness of $\me{\mathbf{S}}{\mathbf{v}}$ in zero-knowledge for any tensor $\mathbf{S}$ (assumed to be one-dimensional for simplicity) and any $\mathbf{v}$ with matching dimensionality:

\begin{itemize}[]
    \item $\tt{pp}\gets \tt{KeyGen}(1^\lambda)$ generates the public parameters used in the scheme, where $\lambda$ is the security parameter of the scheme.
    \item $\com{\mathbf{S}}\gets \tt{Commit}(\mathbf{S}, r; \tt{pp})$ generates a binding and hiding commitment $\com{\mathbf{S}}$ of $\mathbf{S}$, such that $\com{\mathbf{S}}$ leaks no information about $\mathbf{S}$, and no polynomial-time adversary can compute $\mathbf{S}'\neq \mathbf{S}$ and $r'$ such that $\com{\mathbf{S}}= \tt{Commit}(\mathbf{S}; \tt{pp}, r)$. 
    \item $(y, \pi)\gets \tt{ProveEval}(\mathbf{S}, \com{\mathbf{S}}, \mathbf{v}, r; \tt{pp})$ allows the prover to compute $y \gets \me{\mathbf{S}}{\mathbf{\mathbf{v}}}$ for any $\mathbf{v}$ with matching dimensionality, and creates a \emph{proof of evaluation} that $y = \me{\mathbf{S}}{\mathbf{\mathbf{v}}}$ with respect to the committed $\mathbf{S}$. 
    \item $\tt{True/False} \gets \tt{Verify}(y, \pi, \com{\mathbf{S}}, \mathbf{v}; \tt{pp})$ allows the verifier to verify the correctness of $y$, such that 
        \begin{itemize}
            \item \textbf{(Completeness)} if $(y, \pi)= \tt{ProveEval}(\mathbf{S}, \com{\mathbf{S}}, \mathbf{v}, r; \tt{pp})$, then the output is $\tt{True}$.
            \item \textbf{(Soundness)} if $y\neq \me{\mathbf{S}}{\mathbf{v}}$, then the output is \tt{False} with $1-\negl{\lambda}$ probability.
            \item \textbf{(Zero-knowledge)} the verifier learns no information beyond $y = \me{\mathbf{S}}{\mathbf{v}}$.
        \end{itemize}
\end{itemize} In the absence of ambiguity, we omit the randomness $r$ and public parameters $\tt{pp}$ in the subsequent context.

In this study, Hyrax \cite{hyrax}, a variant of the Pedersen commitment \cite{pedersen} that does not require a trusted setup, is used as an instantiation of the polynomial commitment scheme. It operates on a cyclic group $\G$ (typically an elliptic curve), with the hardness assumption of the discrete log problem, and is isomorphic to the addition of $\F$. Hyrax is homomorphic, such that $\tt{Commit}(\mathbf{S}_1, r_1) + \tt{Commit}(\mathbf{S}_2, r_2) = \tt{Commit}(\mathbf{S}_1+\mathbf{S}_2, r_1+r_2)$ for any two tensors $\mathbf{S}_1, \mathbf{S}_2$ and randomness $r_1, r_2$. Hyrax achieves linear complexity in $\tt{Commit}$ and $\tt{ProveEval}$ with respect to the dimensionality $D$ of the tensor $\mathbf{S}$ involved and can be further parallelized by concurrently handling operations on all dimensions. It also balances the commitment size, proof size, and verifier's proof evaluation time, all to sub-linear complexities of $O(\sqrt{D}), O(\log D)$, and $O(\sqrt{D})$ respectively. These improvements are adeptly adopted in zkLLM to minimize both computation and communication burdens.

\subsection{Lookup arguments} \label{sec:lookup}

The lookup argument is commonly used to address non-arithmetic operations within the domain of zero-knowledge proofs \cite{DBLP:journals/ftsec/Thaler22}. Inspired by recent advances \cite{plookup, caulk, caulk+, flookup, baloo, cq}, the incorporation of lookup arguments into zero-knowledge verifiable deep learning inference \cite{zkml} has been pursued. In such a setting, a lookup argument verifies that each element in a secret tensor $\mathbf{S}^D$, known only to the prover, is contained within a predefined table $\mathbf{T}\in \F^N$, mutually acknowledged by both parties. However, the requisite computations for lookup arguments are intrinsically sequential, which contradicts the parallelism preferred in deep learning environments. Furthermore, deploying lookup arguments entails a trade-off between sacrificing precision and incurring excessive memory consumption and trusted setup burdens due to the expansive size of the lookup tables necessary to cover all possible values (with $N$ being significantly large).

In response to these challenges, our proposed lookup arguments for non-arithmetic tensor operations markedly enhance parallelization compared to the largely sequential approaches traditionally used in verifiable deep learning inference. Additionally, our novel proof protocol, tailored for the Softmax function within the attention mechanisms of Transformer models, is designed to optimize the balance between setup and proving times, memory consumptions, and precision.

\subsection{Settings and security assumptions} \label{sec:security-assumptions}

We follow the widely recognized framework for zero-knowledge verifiable inferences as outlined in prior research on zero-knowledge machine learning \cite{zen, zkcnn, ezdps, mystique, vcnn, pvcnn, zkml}. In this framework, the \textbf{prover} (such as an AI company) owns an LLM with a publicly known structure (e.g., described in the technical report), while considering the model's weights as its intellectual property. The prover provides API access to this model for a \textbf{verifier} (like an AI regulation enforcer), who submits a prompt and requests formal proof that the inference result returned by the API is accurate in relation to the prompt and the confidential model.

A semi-honest assumption is applied to the verifier: the verifier accurately reports the outcome of the proof verification (whether it is accepted or rejected) but endeavors to glean additional information about the LLM (like hidden parameters) beyond merely confirming the correctness of the inference result.

In this study, we assume the use of a commitment scheme that ensures \(\lambda\)-bit security. Correspondingly, the computations are carried out within a finite field \(\F\), characterized by a prime order of at least \(\Omega(2^{2\lambda})\). Furthermore, we postulate that every aspect of the transformer model and the data—including the number of layers, the dimensions of tensors, and the complexity of operations between them—is polynomially bounded by \(\lambda\).

\section{\tt{tlookup}: verifiable non-arithmetic operations for deep learning} \label{sec:tlookup}

In this section, we introduce \tt{tlookup}, our novel approach to addressing general non-arithmetic operations in deep learning. The \tt{tlookup} design preserves the widely-used tensor-based structure, guaranteeing seamless compatibility with the established computational frameworks in deep learning. \tt{tlookup} acts as a foundational component of \tt{zkAttn}, our specialized ZKP protocol tailored for the attention mechanism, detailed in Section \ref{sec:zkattn}. Furthermore, \tt{tlookup} is applicable to other non-arithmetic operations essential to the inference mechanisms within LLMs.

We first reduce the non-arithmetic tensor operations to lookup arguments over tensors. Specifically, for a tensor $\mathbf{S} \in \F^D$, the prover aims to convince the verifier that each element of $\mathbf{S}$ exists within $\mathbf{T}\in \F^N$, a table that both parties have full knowledge of. The essence of our approach hinges on the subsequent lemma:

\begin{Lem}[\cite{DBLP:journals/iacr/Habock22a}]
Given tensors $\mathbf{S} \in \F^D$ and $\mathbf{T}\in \F^N$, $\mathbf{S} \subset \mathbf{T}$ as sets if and only if there exists $\mathbf{m}\in \F^N$ such that the following identity of rational functions is satisfied:
\begin{equation}
\sum_{i \in \bracket{D}} \frac{1}{X + \mathbf{S}_i} = \sum_{i \in \bracket{N}} \frac{\mathbf{m}_i}{X + \mathbf{T}_i}. \label{eq:hab22}
\end{equation}
\end{Lem}

When the condition $\mathbf{S}\subset \mathbf{T}$ holds, the prover constructs $\mathbf{m}$ as: \begin{equation}
    \mathbf{m}_i \gets \abs{\cbracket{j: \mathbf{S}_j = \mathbf{T}_i}}, \text{ for } 0\leq i \leq N-1.\label{eq:hab22-coefs}
\end{equation} The verifier can then confirm the equality presented in \eqref{eq:hab22} by randomly choosing $X\gets \beta \sim \F$. By defining: \begin{align}
    \mathbf{A} := \left(\frac{1}{\beta+\mathbf{S}_i}\right)_{i=0}^{D-1}, \mathbf{B} := \left(\frac{1}{\beta+\mathbf{T}_i}\right)_{i=0}^{N-1}, \label{eq:hab22-invs}
\end{align} the aforementioned equality at the random point $\beta$ can be restated as: \begin{equation}
    \sum_{i\in \bracket{D}}\mathbf{A}_i = \sum_{i\in \bracket{N}}\mathbf{m}_i\mathbf{B}_i. \label{eq:hab22-check}
\end{equation} Therefore, with the randomness $\mathbf{u} \sim \F^{\log_2 D}$ and $\alpha \sim \F$, the sumcheck for the correctness of \eqref{eq:hab22-invs} and \eqref{eq:hab22-check} can be formulated as \begin{align}
    0 =& \left(\sum_{\mathbf{i} \in \bracket{D}} \me{\mathbf{A}}{\mathbf{i}} - \sum_{\mathbf{j} \in \bracket{N}}\me{\mathbf{m}}{\mathbf{j}}\me{\mathbf{B}}{\mathbf{j}}\right)\nonumber\\
    &+ \alpha \left(\sum_{\mathbf{i} \in \bracket{D}} \me{\mathbf{e}}{\mathbf{u}, \mathbf{i}} \me{\mathbf{A}}{\mathbf{i}} \left(\me{\mathbf{S}}{\mathbf{i}} + \beta \right) - 1\right)\nonumber\\
    &+ \alpha^2 \left(\sum_{\mathbf{j} \in \bracket{N}} \me{\mathbf{e}}{\mathbf{u}_{\bracket{\log_2\frac{D}{N}:}}, \mathbf{j}} \me{\mathbf{B}}{\mathbf{j}} \left(\me{\mathbf{T}}{\mathbf{j}} + \beta \right) - 1\right), \label{eq:tlookup-sumcheck-preliminary}
\end{align} or equivalently, \begin{align}
    \alpha + \alpha^2 =& \sum_{\mathbf{i}\in \bracket{\frac{D}{N}}}\sum_{\mathbf{j} \in \bracket{N}}\left( \me{\mathbf{A}}{\mathbf{i}\oplus\mathbf{j}}\left(\alpha\me{\mathbf{e}}{\mathbf{u}, \mathbf{i}\oplus\mathbf{j}}\left(\me{\mathbf{S}}{\mathbf{i}\oplus\mathbf{j}} + \beta \right) + 1\right) \right.\nonumber\\&+ \left.ND^{-1}\me{\mathbf{B}}{\mathbf{j}} \left(\alpha^2\me{\mathbf{e}}{\mathbf{u}_{\bracket{\log_2\frac{D}{N}:}}, \mathbf{j}}\left(\me{\mathbf{T}}{\mathbf{j}} + \beta \right)-\me{\mathbf{m}}{\mathbf{j}}\right)\right). \label{eq:tlookup-sumcheck}
\end{align}

A comprehensive description of the procedure to validate $\mathbf{S} \subset \mathbf{T}$ is found in Protocol \ref{protocol:tlookup}. In particular, in Line \ref{alg-line:tlookup-setup}, \Call{tlookup-Setup}{$\mathbf{T}$} generates a short witness $\dbbracket{\mathbf{T}}$ to a prescribed table $\mathbf{T}$ known to both parties; in Line \ref{alg-line:tlookup-prep}, the prover constructs $\mathbf{m}$ based on a tensor $\mathbf{S}$ and table $\mathbf{T}$ and commit to $\mathbf{S}$ and $\mathbf{m}$ using \Call{tlookup-Prep}{$\mathbf{S}, \mathbf{T}$}; finally, in Line \ref{alg-line:tlookup-prove}, $\innerproduct{\mathcal{P}}{\mathcal{V}}$.\Call{tlookup-Prove}{$\dbbracket{\mathbf{S}}, \dbbracket{\mathbf{m}}, \dbbracket{\mathbf{T}}$} is the interactive process of the prover $\mathcal{P}$ proving that a secret tensor $\mathbf{S}$ is elementwisely in $\mathbf{T}$, which has been committed as $\dbbracket{\mathbf{T}}$.

\begin{protocol}
    \caption{\tt{tlookup}}
    \label{protocol:tlookup}
    \begin{algorithmic}[1]
        \Require The prover $\mathcal{P}$ knows $\mathbf{S}\in \F^D$. $N, D$ are both powers of 2 such that $N$ divides $D$. 

        \Procedure{tlookup-Setup}{$\mathbf{T}\in \F^N$} \label{alg-line:tlookup-setup} 
            \State \Return $\dbbracket{\mathbf{T}} \gets \tt{Commit}(\mathbf{T}; 0)$ \Comment{No hiding required}
        \EndProcedure

        \Procedure{$\mathcal{P}$.tlookup-Prep}{$\mathbf{S}\in \F^D, \mathbf{T}\in \F^N$} \label{alg-line:tlookup-prep} 
            \State Compute $\mathbf{m} = \mathbf{m}(\mathbf{S}, \mathbf{T})$ as \eqref{eq:hab22-coefs}
            \State $\mathcal{P}\to \mathcal{V}: \dbbracket{\mathbf{S}} \gets \tt{Commit}(\mathbf{S})$ \label{alg-line:tlookup-commit}
            \State $\mathcal{P}\to \mathcal{V}: \dbbracket{\mathbf{m}} \gets \tt{Commit}(\mathbf{m})$ 
        \EndProcedure

        \Procedure{$\innerproduct{\mathcal{P}}{\mathcal{V}}$.tlookup-Prove}{$\dbbracket{\mathbf{S}}$, $\dbbracket{\mathbf{m}}, \dbbracket{\mathbf{T}}$} \label{alg-line:tlookup-prove}
            \State $\mathcal{V}\to \mathcal{P}: \beta \sim \F$ \label{alg-line:tlookup-random-challenge}
            \State $\mathcal{P}$ computes $\mathbf{A}, \mathbf{B}$ as \eqref{eq:hab22-invs}
            \State $\mathcal{P}\to \mathcal{V}: \dbbracket{\mathbf{A}}\gets \tt{Commit}(\mathbf{A}), \dbbracket{\mathbf{B}}\gets \tt{Commit}(\mathbf{B})$
            \State $\mathcal{P}$ and $\mathcal{V}$ run the sumcheck on \eqref{eq:tlookup-sumcheck}, followed by the proofs of evaluation on $\dbbracket{\mathbf{A}}, \dbbracket{\mathbf{B}}, \dbbracket{\mathbf{S}}, \dbbracket{\mathbf{m}}$ and $\dbbracket{\mathbf{T}}$. \label{alg-line:tlookup-sumchecks}
        \EndProcedure

    \end{algorithmic}
\end{protocol}

Meanwhile, for elementwise non-arithmetic operations \(f : \mathcal{X} \to \mathcal{Y}\) over tensors where \(\mathcal{X}, \mathcal{Y}\subset \F\), two lookup tables can be constructed: \(\mathbf{T}_\mathcal{X} := \left(x\right)_{x\in \mathcal{X}}\) and \(\mathbf{T}_\mathcal{Y} := \left(f(x)\right)_{x\in \mathcal{X}}\). To demonstrate that \(\mathbf{Y} = f(\mathbf{X})\) for some \(\mathbf{X}, \mathbf{Y}\in \F^D\) (broadcasting \(f\) over all dimensions), one can apply the idea of random linear combination to reduce the check to one instance of Protocol \ref{protocol:tlookup} where \(\mathbf{X} + \alpha \mathbf{Y}\subset \mathbf{T}_\mathcal{X} + \alpha \mathbf{T}_\mathcal{Y}\) for \(\alpha\sim\F\) chosen by the verifier. 

\begin{Exm}[ReLU with rescaling]\label{exm:relu}
    We first consider the rectified linear unit (ReLU), which is a common activation function in contemporary deep learning models, including Transformers. ReLUs are generally applied subsequent to linear layers (for instance, fully connected layers) where products are involved. In the scenario of fully quantized computation, it becomes necessary for the ReLU to incorporate rescaling as well. This is denoted as follows:
    \begin{equation}
        \mathbf{A} \gets \text{ReLU}(\mathbf{Z}) = \round{\frac{\mathbf{Z}}{\gamma}}\odot \1\left\{\round{\frac{\mathbf{Z}}{\gamma}} \geq 0\right\},
    \end{equation}
    where \(\gamma\) represents the scaling factor used in the system (assumed to be even for simplicity). We assume that \(-\frac{B}{2} \leq \round{\frac{\mathbf{Z}}{\gamma}} < \frac{B}{2}\) holds element-wise for a positive even integer \(B\). Considering that \(\mathbf{Z}\) is decomposed as \(\mathbf{Z}' := \round{\frac{\mathbf{Z}}{\gamma}}\) and \(\mathbf{R} = \mathbf{Z} - \gamma \mathbf{Z}'\), we establish a pair of input-output lookup tables for \(\mathbf{Z}'\). These are defined as \(\mathbf{T}_\mathcal{X}:= \left[-\frac{B}{2}, \frac{B}{2}-1\right]\) and \(\mathbf{T}_\mathcal{Y}:= \mathbf{T}_\mathcal{X}^+\) (i.e., taking the maximum with 0 element-wise), and an additional lookup table for \(\mathbf{R}\) as \(\mathbf{T}_\mathcal{R} = \left[-\frac{\gamma}{2}, \frac{\gamma}{2} - 1\right]\). By requiring the prover to demonstrate to the verifier that \(\mathbf{Z}'+\alpha \mathbf{A} \subset \mathbf{T}_\mathcal{X} + \alpha \mathbf{T}_\mathcal{Y}\) for a random \(\alpha\), and that \(\mathbf{R} \subset \mathbf{T}_\mathcal{R}\), both using Protocol \ref{protocol:tlookup}, in addition to proving the decomposition as \(\mathbf{Z} = \gamma \mathbf{Z}' + \mathbf{R}\), we can sufficiently validate the correctness of inference through the ReLU function. Notably, unlike the brute-force method that employs a single lookup table and incurs an \(O(B \gamma)\) overhead in both running time and memory usage, the use of two lookup tables effectively reduces this overhead to \(O(B + \gamma)\). Similarly, if \(\gamma\) is too large to fit the table into memory, it can be further divided into a \(K\)-digit \(\gamma^{\frac{1}{K}}\)-ary number. In this scenario, each of the \(K\) digits in the remainder corresponds to a separate \texttt{tlookup}, thus adequately covering all possible values of the remainder.
\end{Exm}

However, resolving the long-standing problem of excessive memory consumption for lookup tables in the realm of deep learning requires additional efforts. Specifically, in Section \ref{sec:zkattn}, \tt{tlookup} is further refined into \tt{zkAttn} to address its multivariate and highly non-arithmetic nature, optimizing the balance among running time, memory consumption, and approximation error.

\section{\tt{zkAttn}: dedicated ZKP for the attention mechanism in LLMs} \label{sec:zkattn}

The attention mechanism is a key component in modern transformers, including state-of-the-art LLMs. However, incorporating these mechanisms into ZKP backends has been challenging, primarily due to their distinctive mathematical properties. Specifically, the Softmax function, integral to the attention mechanism, involves non-arithmetic operations like exponentiation, and its multivariate aspect complicates the use of polynomial approximations for traditional ZKP backends. To address these challenges, we introduce \texttt{zkAttn}, a specialized ZKP tailored for the attention mechanism, designed to leverage its inherent mathematical characteristics effectively.

\subsection{Formulation of \tt{zkAttn}} \label{sec:zkattn-formulation}

The attention mechanism, in its discretized form, accepts as input a value matrix $\mathbf{V}\in \F^{n\times d}$, a key matrix $\mathbf{K}\in \F^{n\times d}$, and a query matrix $\mathbf{Q} \in \F^{m\times d}$. It produces the output $\text{Softmax}\left(\frac{\mathbf{Q} \mathbf{K}^\top}{\sqrt{d}}\right)\mathbf{V}$ subject to appropriate rescaling of the input and output due to quantization, where Softmax is applied row-wise. In this discussion, we focus on $\text{Attention}(\mathbf{Q}, \mathbf{K}, \mathbf{V}) = \text{Softmax}\left(\frac{\mathbf{Z}}{\sqrt{d}}\right)$, where the input matrix $\mathbf{Z} = \mathbf{Q}\mathbf{K}^\top$ is presumed to be scaled by the scaling factor $\gamma$ from its actual values. It is assumed that $d$ is a constant, known to both the prover and verifier, stemming from the presumption of a known model architecture. Equivalently, for each row $\mathbf{z} = (z_0, z_1, \dots, z_{n-1}) \in \F^n$, the objective is to devise an algorithm that computes \begin{equation} \label{eq:softmax}
    s(\mathbf{z}) := \left(\frac{\exp\left(\frac{z_i}{\gamma\sqrt{d}}\right)}{\sum_{j=0}^{n-1} \exp\left(\frac{z_j}{\gamma\sqrt{d}}\right)}\right)_{i=0}^{n-1}
\end{equation} in the real domain. Alternatively, it should compute its quantized counterpart $\theta s(\mathbf{z})$, ensuring limited numerical error and manageable proof generation overhead. Here, $\theta$ represents the scaling factor of all Softmax outputs in the system. Notably, this factor differs from the scaling factor $\gamma$ used for other matrices, such as $\mathbf{Q}, \mathbf{K}$, and $\mathbf{V}$. For the sake of streamlined and verifiable rescaling in subsequent computations, we posit that $\theta$ is a multiple of $\gamma$.

To circumvent the verification of real division operations—which can lead to remainders after quantization—we observe the following. By utilizing the shift-invariance property of Softmax and defining \begin{equation} 
\label{eq:shift-const}
    \hat{z} := \gamma\sqrt{d} \ln \left(\sum_{j=0}^{n-1}\exp\left(\frac{z_i}{\gamma\sqrt{d}}\right)\right),
\end{equation} we derive that \begin{align}
    s(\mathbf{z}) &= s(\mathbf{z} - \hat{z})
    = \left(\exp\left(\frac{z_i - \hat{z}}{\gamma\sqrt{d}}\right)\right)_{i=0}^{n-1}.
\end{align} It is imperative to understand that the computation of $\hat{z}$ ($\round{\hat{z}}$ to be specific, since $\hat{z}$ is not an integer in general) in \eqref{eq:shift-const} is not directly verified due to its highly non-arithmetic nature. Instead, the prover ensures that the output of \(s(\mathbf{z})\) adheres to proper normalization. In its quantized representation, the sum of its dimensions must equal \(\theta\). A certain degree of deviation is acceptable owing to quantization, and the precise bounds of this error will be elucidated in Section \ref{sec:error-anal}. Furthermore, beyond verifying normalization, there exists the challenge of crafting a scheme to compute the quantized exponentiation. This scheme should not only be accurate, approximating \(\theta \exp\left(\frac{\cdot}{\gamma\sqrt{d}}\right)\) with minimal error, but also be amenable to efficient verification through the proof protocol that will be subsequently introduced.

Observe that, given the definition of $\hat{z}$, $z_i - \hat{z} \leq 0$ for all $i$s. On the other hand, as $z_i$s are all real numbers involved in the matrix multiplication scaled by $\gamma$, it is also reasonable to assume that each $z_i-\hat{z}$ is lower bounded by some integer $-B$ such that $\gamma \ll B \ll |\F|$, such that $(-B, 0]$ can accommodate the reasonable values of $z_i - \hat{z}$, but sufficiently small so as not to cause wraparounds in $\F$. 


Without loss of generality, consider \(B\) as a product of \(K\) positive integers, denoted as \(B = \prod_{k=0}^{K-1} b^{(k)}\). A bijection can then be established between \(\bracket{B}\) and the product space \(\prod_{k=0}^{K-1}\bracket{b^{(k)}}\). By defining \(B^{(k)}\) as \[
B^{(k)}:=\begin{cases} 
1, & \text{if } k = 0; \\
\prod_{j=0}^{k-1}b^{(j)}, & 1 \leq k \leq K-1,
\end{cases}
\]
our bijection $\mathfrak{b} : \prod_{k=0}^{K-1}\bracket{b^{(k)}} \to [B]$ can be expressed as 
\begin{equation} \label{eq:isomorphism}
    \mathfrak{b}\left(x^{(0)}, x^{(1)}, \dots, x^{(K-1)}\right) = \sum_{k=0}^{K-1} x^{(k)} B^{(k)}.
\end{equation}
Consequently, for each \(\left(x^{(0)}, x^{(1)}, \dots, x^{(K-1)}\right) = \mathfrak{b}^{-1}(x)\), the following holds: 
\begin{align} \label{eq:decomp}
     \exp\left(-\frac{x}{\gamma 
    \sqrt{d}}\right)
    &=  \exp\left(-\frac{\sum_{k=0}^{K-1} x^{(k)} B^{(k)}}{\gamma 
    \sqrt{d}}\right) = \prod_{k=0}^{K-1} \exp\left(-\frac{B^{(k)}}{\gamma\sqrt{d}}x^{(k)}\right). 
\end{align}
Our objective is to compute the quantized representation of equation \eqref{eq:decomp}, taking into account the scaling factor \(\gamma\). If we further decompose \(\gamma\) as \(\theta = \prod_{k=0}^{K-1}\theta^{(k)}\) with non-negative values for \(\theta^{(k)}\), equation \eqref{eq:decomp} gives rise to 
\begin{align} \label{eq:decomp-rescaled}
    \theta \exp\left(-\frac{x}{\gamma 
    \sqrt{d}}\right) &= \prod_{k=0}^{K-1}\theta^{(k)} \exp\left(-\frac{B^{(k)}}{\gamma\sqrt{d}}x^{(k)}\right).
\end{align} 

Following the decomposition in Equation \eqref{eq:decomp-rescaled}, we can construct \( K \) \tt{tlookup} tables \( \mathbf{T}^{(k)} = \left(\mathbf{T}^{(k)}_\mathcal{X}, \mathbf{T}^{(k)}_\mathcal{Y}\right) \). Each table \( \mathbf{T}^{(k)} \) comprises all potential input-output pairs corresponding to the \( k \)-th term in the product of \eqref{eq:decomp-rescaled}: \begin{align} 
\label{eq:lookup}
    \mathbf{T}_\mathcal{X}^{(k)} := \bracket{b^{(k)}},\quad \mathbf{T}_\mathcal{Y}^{(k)} := \left(\round{\theta^{(k)}\exp\left(-\frac{B^{(k)}}{\gamma\sqrt{d}}x\right)}\right)_{x\in \bracket{b^{(k)}}}.
\end{align} Given any input \( z \in (-B, 0] \), the prover first decomposes \( \mathfrak{b}(-z) = \left(x^{(0)}, x^{(1)}, \dots, x^{(K-1)}\right) \) according to \eqref{eq:isomorphism}. Each component \( x^{(k)} \) is subsequently mapped to \( y^{(k)} \) based on \( \mathbf{T}^{(k)} \), resulting in the computation \( y\gets \prod_{k=0}^{K-1}y^{(k)} \). Subsequently, the prover must demonstrate to the verifier that: 

\begin{align}
    z + \sum_{k=0}^{K-1} x^{(k)}B^{(k)} &= 0, & ~ \label{eq:valid-decomp}\\
    \left(x^{(k)}, y^{(k)}\right) &\in \mathbf{T}^{(k)}, & \forall 0\leq k \leq K-1, \label{eq:valid-lookup}\\
    \prod_{k=0}^{K-1} y^{(k)} &= y. \label{eq:valid-prod} 
\end{align}

Equation \eqref{eq:valid-decomp} confirms that the decomposition of \( -z \) is valid. Equation \eqref{eq:valid-lookup} ensures the correctness of the exponent in each component concerning the pre-computed values in \( \mathbf{T}^{(k)} \), and \eqref{eq:valid-prod} asserts that the output \( y \) is accurately derived using the homomorphism of exponentiation from each factor \( y^{(k)} \). Together, these three conditions guarantee the correct computation of the exponentiation operation, up to the rounding errors. Specifically:

\begin{Lem}
    Conditions \eqref{eq:valid-decomp}, \eqref{eq:valid-lookup}, and \eqref{eq:valid-prod} imply:
    \begin{equation}
        y = \prod_{k=0}^{K-1}\round{\theta^{(k)}\exp\left(-\frac{B^{(k)}}{\gamma\sqrt{d}}x^{(k)}\right)},
    \end{equation}
    where \( {\left(x^{(k)}\right)}_{k=0}^{K-1} = \mathfrak{b}(-z)\) is the valid decomposition according to \eqref{eq:isomorphism}.
\end{Lem}

The deviation between \( y \) and the exact scaled exponent \[ \theta\exp\left(\frac{z}{\gamma\sqrt{d}}\right) = \prod_{k=0}^{K-1}\theta^{(k)}\exp\left(-\frac{B^{(k)}}{\gamma\sqrt{d}}x^{(k)}\right) \] arises only from the rounding of each factor. An in-depth analysis of this will be covered in Section \ref{sec:error-anal}, where we will also provide guidance on selecting the parameters \( \theta^{(k)} \) and \( B^{(k)} \). In the subsequent sections of this text, we delve into the protocol design, facilitating the batched verification of \eqref{eq:valid-decomp}, \eqref{eq:valid-lookup}, and \eqref{eq:valid-prod} for each dimension of large tensors used in transformer computations.

\subsubsection{Optimization for the most and least significant segments}\label{sec:zkattn-optimization} For the uppermost significant $M$ segments, specifically $x^{(k)}$ with $K-M \leq k \leq K - 1$, consider a scenario where if any of these segments $x^{(k)}$ have non-zero values, the resulting exponent \begin{equation}
    \exp\left(-\frac{x}{\gamma\sqrt{d}}\right) \leq \exp\left(-\frac{B_{K-M}}{\gamma\sqrt{d}}\right)
\end{equation} approximates 0 closely enough that the output $y$ from \eqref{eq:valid-prod} can be designated as 0. This outcome can be achieved by configuring each table $\mathbf{T}^{(k)}$ that $K-M \leq k \leq K-1$ in \eqref{eq:valid-lookup}, to yield $y^{(k)} = 0$ for any $x^{(k)} > 0$. Moreover, based on our initial design, instances where $x^{(k)} = 0$, the value of $y^{(k)}$ defaults to $\round{\theta^{(k)}}$. Clearly, assigning any value other than $1$ to these $\theta^{(k)}$ would only amplify the errors in $\mathbf{T}^{(k)}$ and other tables, especially under the constraint that $\prod_{k=0}^{K-1}\theta^{(k)} = \theta$ is constant. Therefore, for these most significant $M$ segments, the lookup tables $\mathbf{T}^{(k)}$ can be reduced to the indicator function $y^{(k)} = \1\{x^{(k)} = 0\}$.

On the other hand, for the least significant $L$ segments $x^{(k)}$, indexed by $0 \leq k \leq L-1$, the expression $\exp\left(-\frac{B^{(k)}}{\gamma\sqrt{d}}x^{(k)}\right)$ tends to hover close to 1 for all possible values of $0 \leq x^{(k)} \leq b^{(k)} - 1$. Given this, approximating the exponentiation as a constant of 1 incurs a negligible error. Analogous to the strategy for the most significant segments, it is efficient to set the scaling factors $\theta^{(k)}$ to 1, sidestepping larger alternatives. This approach frees up room for allocating larger $\theta^{(k)}$s for segments indexed by $L \leq k \leq K - M -1$, thereby enhancing precision for these segments. As a result, the constraint \eqref{eq:valid-lookup} for validating input-output pairs simplifies to $x^{(k)} \in \bracket{b^{(k)}}$, given that $y^{(k)}$ consistently equates to $1$.

As delineated in Section \ref{sec:error-anal}, employing these optimizations for both the most and least significant segments tightly upper bounds the error in \tt{zkAttn}, aligning closely with the computational logic of the original neural networks.

\subsection{\tt{zkAttn}: the main protocol}

In Protocol \ref{protocol:zkattn}, we present the technical details of \texttt{zkAttn} built upon \texttt{tlookup}, our protocol for general non-arithmetic operations in deep learning. \texttt{zkAttn} is separated into three steps. First, \Call{zkAttn-Setup}{$\cdot$} (in Line \ref{alg-line:zkattn-setup}) completes the setup for \texttt{zkAttn} by generating short witnesses of all tables involved. Then, in \Call{zkAttn-Compute}{$\cdot$}, the prover is responsible for computing the output of the attention mechanism and all necessary auxiliary tensors (e.g., the normalization constants, the inputs and outputs of each segment, and the recovered row-wise sum that should not excessively deviate from 1) for the zero-knowledge verifiability of the attention mechanism, and sends the commitments of these tensors to the verifier. Finally, the prover and verifier engage in the interactive protocol \Call{zkAttn-Prove}{$\cdot$} in Line \ref{alg-line:zkattn-compute}, which involves proving the correctness of each segment and the normalization using the specialized lookup arguments of \texttt{tlookup}, as well as all arithmetic relations connecting the auxiliary tensors.

\begin{protocol*}
    \caption{\tt{zkAttn} (See Section \ref{sec:taxonomy-matmul} about the two matrix multiplications involved)}
    \label{protocol:zkattn}
    \begin{algorithmic}[1]
        \Require Both the prover $\mathcal{P}$ and the verifier $\mathcal{V}$ know: the lower bound of input $-B$, and the factorization $B = \prod_{k=0}^{K-1}b^{(k)}$; the number of the most and least significant segments $M$ and $L$ in Section \ref{sec:zkattn-optimization}; the scaling factor of the input $\gamma$, the output $\theta$, and each segment $\theta^{(k)}$ for each $L \leq k \leq K-M-1$; the parameters $m, n, d$ related to the dimensions of the input; the tolerable error $E$ in row-wise normalization. 

        \Procedure{zkAttn-Setup}{$B, K, M, L, \left(b^{(k)}\right)_{k=0}^{K-1}, \gamma, \theta, \left(\theta_K\right)_{k=L}^{K-M-1}, m, n, d, E$} \label{alg-line:zkattn-setup}
            \For{$0\leq k \leq K-1$} 
                \State $\dbbracket{\mathbf{T}_\mathcal{X}^{(k)}}\gets \Call{tlookup-Setup}{\mathbf{T}_\mathcal{X}^{(k)}}$  \Comment{$\mathbf{T}_\mathcal{X}^{(k)} = \bracket{b^{(k)}}$, i.e., the input of the $k$-th segment}
            \EndFor

            \For{$L \leq k \leq K - M - 1$} 
                \State $\dbbracket{\mathbf{T}_\mathcal{Y}^{(k)}}\gets \Call{tlookup-Setup}{\mathbf{T}_\mathcal{Y}^{(k)}}$ \Comment{$\mathbf{T}_\mathcal{Y}^{(k)}$ as defined in \eqref{eq:lookup}, i.e., the output of the $k$-th segment}
            \EndFor

            \For{$K-M \leq k \leq K - 1$}  
                \State $\dbbracket{\mathbf{T}_\mathcal{Y}^{(k)}}\gets \Call{tlookup-Setup}{\mathbf{T}_\mathcal{Y}^{(k)}}$ \Comment{$\mathbf{T}_\mathcal{Y}^{(k)} = \1\left\{\bracket{b^{(k)}} = 0\right\}$, i.e., the optimized output of the $k$-th segment}
            \EndFor

            \State $\dbbracket{\mathbf{T}_\mathcal{R}}\gets \Call{tlookup-Setup}{\mathbf{T}_\mathcal{R}}$ \Comment{$\mathbf{T}_\mathcal{R}= [\theta - E, \theta + E]$, i.e., all tolerable values of row-wise sum of the output}

            \State \Return $\left(\dbbracket{\mathbf{T}_\mathcal{X}^{(k)}}\right)_{k=0}^{K-1}, \left(\dbbracket{\mathbf{T}_\mathcal{Y}^{(k)}}\right)_{k=L}^{K-1}, \dbbracket{\mathbf{T}_\mathcal{R}}$
        \EndProcedure

        \Procedure{$\mathcal{P}$.zkAttn-Compute}{$\mathbf{Z} \in \F^{m\times n}$, $\left(\mathbf{T}_\mathcal{X}^{(k)}\right)_{k=0}^{K-1}, \left(\mathbf{T}_\mathcal{Y}^{(k)}\right)_{k=L}^{K-1}$} \Comment{Some implicit parameters included in \Call{Setup}{$\cdot$} omitted} \label{alg-line:zkattn-compute}
            \State $\mathbf{Z}' \gets \mathbf{Z} - \round{\Hat{\mathbf{z}}}\mathbf{1}^\top$, where $\Hat{\mathbf{z}}\in \mathbb{R}^m$ is computed row-wise as \eqref{eq:shift-const} 
            
            \State $\left(\mathbf{X}^{(k)}\right)_{k=0}^{K-1}\gets \mathfrak{b}^{-1}\left(-\mathbf{Z}'\right)$ \Comment{$-\mathbf{Z}$ is decomposed elementwisely}
            
            \For{$k\gets L, L+1, \dots, K-1$}
                \State $\mathbf{Y}^{(k)} \gets f^{(k)}(\mathbf{X}^{(k)})$ elementwisely, where $f^{(k)}$ is defined by  $\mathbf{T}_\mathcal{X}^{(k)}, \mathbf{T}_\mathcal{Y}^{(k)}$
                
            \EndFor
            \State $\mathbf{Y}\gets \bigodot_{k=L}^{K-1}\mathbf{Y}^{(k)}$ \Comment{Compute the final output}
            \State $\Hat{\mathbf{y}}\gets \mathbf{Y}.\text{sum(axis = 1)}$ \Comment{For checking the normalization of each row}
            \State $\mathcal{P}\to \mathcal{V}: \dbbracket{\mathbf{Z}}, \dbbracket{\round{\Hat{\mathbf{z}}}}, \left(\dbbracket{\mathbf{X}^{(k)}}\right)_{k=0}^{K-1}, \dbbracket{\mathbf{Y}}, \left(\dbbracket{\mathbf{Y}^{(k)}}\right)_{k=L}^{K-1}, \dbbracket{\Hat{\mathbf{y}}}$
        \EndProcedure

        \Procedure{$\innerproduct{\mathcal{P}}{\mathcal{V}}$.zkAttn-Prove}{$\dbbracket{\mathbf{Z}}, \dbbracket{\round{\Hat{\mathbf{z}}}}, \left(\dbbracket{\mathbf{X}^{(k)}}\right)_{k=0}^{K-1}, \dbbracket{\mathbf{Y}}, \left(\dbbracket{\mathbf{Y}^{(k)}}\right)_{k=L}^{K-1}, \dbbracket{\Hat{\mathbf{y}}}$} \label{alg-line:zkattn-prove}
            \For{$k\gets 0, 1, \dots, K-1$}
                \State $\mathcal{P}$.\Call{tlookup-Prep}{$\mathbf{X}^{(k)}, \mathbf{T}_\mathcal{X}^{(k)}$} \Comment{$\dbbracket{\mathbf{m}^{(k)}:=\mathbf{m}\left(\mathbf{X}^{(k)}, \mathbf{T}_\mathcal{X}^{(k)}\right)}$ transmitted to $\mathcal{V}$}
                \State \Call{$\innerproduct{\mathcal{P}}{\mathcal{V}}$.tlookup-Prove}{$\dbbracket{\mathbf{X}^{(k)}}$, $\dbbracket{\mathbf{m}^{(k)}}, \dbbracket{\mathbf{T}_\mathcal{X}^{(k)}}$} \Comment{Prove the correctness on the $k$-th segment}
            \EndFor
            \State $\mathcal{V}\to \mathcal{P}: \alpha\sim\F$
            \For{$k \gets L, L+1, \dots K-1$}
                \State $\mathcal{P}$.\Call{tlookup-Prep}{$\mathbf{X}^{(k)} + \alpha \mathbf{Y}^{(k)}, \mathbf{T}_\mathcal{X}^{(k)} + \alpha \mathbf{T}_\mathcal{Y}^{(k)}$} \Comment{$\dbbracket{\mathbf{m}^{(k)}:=\mathbf{m}\left(\mathbf{X}^{(k)} + \alpha \mathbf{Y}^{(k)}, \mathbf{T}_\mathcal{X}^{(k)} + \alpha \mathbf{T}_\mathcal{Y}^{(k)}\right)}$ transmitted to $\mathcal{V}$}
                \State \Call{$\innerproduct{\mathcal{P}}{\mathcal{V}}$.tlookup-Prove}{$\dbbracket{\mathbf{X}^{(k)}} + \alpha\dbbracket{\mathbf{Y}^{(k)}}$, $\dbbracket{\mathbf{m}^{(k)}}, \dbbracket{\mathbf{T}_\mathcal{X}^{(k)}} + \alpha \dbbracket{\mathbf{T}_\mathcal{Y}^{(k)}}$} \Comment{Prove the correctness on the $k$-th segment}
            \EndFor
            \State $\mathcal{P}$.\Call{tlookup-Prep}{$\Hat{\mathbf{y}}, \mathbf{T}_\mathcal{R}$} \Comment{$\dbbracket{\mathbf{m}_\mathcal{R}:=\mathbf{m}\left(\mathbf{Y}, \mathbf{T}_\mathcal{R}\right)}$ transmitted to $\mathcal{V}$}
            \State \Call{$\innerproduct{\mathcal{P}}{\mathcal{V}}$.tlookup-Prove}{$\dbbracket{\Hat{\mathbf{y}}}$, $\dbbracket{\mathbf{m}_\mathcal{R}}, \dbbracket{\mathbf{T}_\mathcal{R}}$} \Comment{Prove the correctness on the $k$-th segment}
            
            \State $\mathcal{P}$ and $\mathcal{V}$ run the sumcheck for $\mathfrak{b}(\mathbf{X}_0, \mathbf{X}_1, \dots, \mathbf{X}_{K-1}) + \mathbf{Z}' = \mathbf{0} \bigwedge \mathbf{Z}' = \mathbf{Z} - \round{\Hat{\mathbf{z}}}\mathbf{1}^\top \bigwedge \mathbf{Y} = \bigodot_{k=L}^{K-1}\mathbf{Y}^{(k)} \bigwedge \Hat{\mathbf{y}} = \mathbf{Y}.\text{sum(axis = 1)}$, followed by the proof of evaluations on $\dbbracket{\mathbf{Z}}, \dbbracket{\round{\Hat{\mathbf{z}}}}, \left(\dbbracket{\mathbf{X}^{(k)}}\right)_{k=0}^{K-1}, \dbbracket{\mathbf{Y}}, \left(\dbbracket{\mathbf{Y}^{(k)}}\right)_{k=L}^{K-1}, \dbbracket{\Hat{\mathbf{y}}}$
        \EndProcedure
    \end{algorithmic}
\end{protocol*}

\section{Putting everything together} \label{pet}

\subsection{Taxonomy of verifiable tensor operations} \label{sec:taxonomy}

In this section, we present a taxonomy of the tensor operations involved in modern LLMs and the customized handling of these operations by zkLLM.

\subsubsection{Matrix multiplications} \label{sec:taxonomy-matmul}
Matrix multiplication plays a crucial role in modern transformers, including all linear layers and positional encoding (e.g., RoPE \cite{rope}) in LLMs.

Dedicated sumchecks \cite{safetynets} for matrix multiplications have achieved running times significantly lower than the computation itself. This development has been a key driver in the creation of specialized ZKPs for deep learning, a method also applied in this study to establish the correctness of all matrix products. To confirm \(\mathbf{C} = \mathbf{A}\mathbf{B}\), with \(\mathbf{A}\in \F^{m\times n}\) and \(\mathbf{B} \in \F^{n\times p}\), the prover and the verifier execute a sumcheck on 
\begin{equation}
    \me{\mathbf{C}}{\mathbf{u}, \mathbf{v}} = \sum_{\mathbf{i}\in \{0, 1\}^{\ceil{\log_2 n}}}\me{\mathbf{A}}{\mathbf{u}, \mathbf{i}}\me{\mathbf{B}}{\mathbf{i}, \mathbf{v}},
\end{equation} 
where \(\mathbf{u}\in \F^{\ceil{\log_2 m}}\) and \(\mathbf{v}\in \F^{\ceil{\log_2 p}}\) are selected at random by the verifier. This specialized proof for matrix multiplication ensures a prover time of $O(mn + np)$, faster than the computation process.

\subsubsection{Activation functions}
To enhance performance, modern LLMs have replaced traditional ReLU activation functions with smoother alternatives like SwiGLU \cite{swiglu} and GELU \cite{gelu}. This transition necessitates extra efforts to make these new activation functions verifiable. The SwiGLU function, parameterized by \(\beta\), is defined as 
\begin{equation}
    \text{SwiGLU}_\beta(z, z') := \text{Swish}_\beta(z) \cdot z',
\end{equation} 
with the Swish function being \(\text{Swish}_\beta(z):= z\cdot \text{Sigmoid}(\beta z)\). Though the non-arithmetic Sigmoid function can be integrated into the proof system via \texttt{tLookup}, optimizing setup costs and memory usage is crucial. This is achieved by reducing the Sigmoid function to \texttt{zkAttn}, given \(\text{Softmax}\begin{pmatrix}z\\ 0\end{pmatrix} = \begin{pmatrix}\text{Sigmoid}(z)\\ \text{Sigmoid}(-z)\end{pmatrix}\), thus circumventing the bottleneck of iterating over extensive input-output pairs. The GELU function, defined as \(\text{GELU}(z) := z\Phi(z) \approx z\text{Sigmoid}(1.702z)\), is handled similarly.

\subsubsection{Normalization}
LLMs employ LayerNorm \cite{layernorm} and its variants (e.g., RMSNorm \cite{rmsnorm}) for training stability. Unlike batch normalization, which can be merged into preceding layers for verifiable inference, LayerNorm involves non-linear transformations within each sample \(\mathbf{x}\), described as 
\begin{equation}
    y \gets \frac{\mathbf{x} - \mathbb{E}[\mathbf{x}]}{\sqrt{\text{Var}{[x]} + \epsilon}}.
\end{equation} 
The compound non-arithmetic operations of square-root and inverse are managed through two sequential \texttt{tlookup} steps. These steps are responsible for the verifiable downscaling of the input and the quantized compound operation, respectively. Similar to the ReLU example presented in Example \ref{exm:relu}, the implementation of the first \texttt{tlookup} for downscaling is designed to reduce the overall sizes of the lookup tables, thereby keeping memory usage at a reasonable level.

\subsection{Assembly of the proofs}

Following the pioneering works \cite{safetynets, zkcnn}, zkLLM utilizes sumcheck-based proofs for computations across various components of LLMs. These proofs are assembled in reverse logical order of the arithmetic circuits, methodically reducing the claimed multilinear extension values of the output \( y \) to those associated with the prompt \( \mathbf{X} \) and the model parameters \( \mathbf{W} \). These are then verified straightforwardly and through proof of evaluations on the commitment \( \dbbracket{\mathbf{W}} \). For high-level clarity, zkLLM can be distilled into three main components, with additional commitments caused by \texttt{tlookup}s omitted for simplicity:

\begin{itemize}
    \item \( \com{\mathbf{W}} \gets \texttt{zkLLM-Commit}(\mathbf{W}, \texttt{pp}, r) \): The prover commits to the model parameters \( \mathbf{W} \) using the public parameters (generators of the commitment scheme) \texttt{pp} and randomness \( r \).
    \item \( (y, \pi) \gets \texttt{zkLLM-Prove}(\mathbf{W}, \mathbf{X}, \texttt{pp}, r) \): The prover computes the output \( y \) with the prompt \( \mathbf{X} \) and model \( \mathbf{W} \), and assembles the proof \( \pi \) using the sumcheck protocols and proofs of evaluations as previously described.
    \item \( b \gets \texttt{zkLLM-Verify}(\mathbf{X}, y, \com{\mathbf{W}}) \): The verifier checks the correctness of each sumcheck and proof of evaluation within \( \pi \), outputting \( b=1 \) to accept the proof (if all components are correctly verified), and \( b=0 \) otherwise.
\end{itemize}

\section{Analysis} \label{sec:analysis}

\subsection{Error analysis on \tt{zkAttn}} \label{sec:error-anal}

In this section, we examine the error introduced by \tt{zkAttn}, as discussed in Section \ref{sec:zkattn}. Our analysis serves three primary objectives: first, to establish an upper bound on the error, thereby demonstrating that our \tt{zkAttn} design remains faithful to the original neural network; second, to fine-tune the parameters integral to \tt{zkAttn}, aiming to minimize the error; and third, to determine an acceptable upper bound on the error upon the verification of proper normalization in \tt{zkAttn}.

The overall error of \tt{zkAttn} originates from two sources, the rounding of the shifting factor $\hat{z}$ in \eqref{eq:shift-const} that makes the normalization no longer perfect, and the rounding of each segment encoded in the lookup tables $\mathbf{T}^{(k)}$s that introduces errors to the exponentiations. The bound of the overall error is stated in Theorem \ref{thm:error-bound} and analyzed in details in Appendix \ref{appendix:error-anal}.

\begin{Thm}[Error bound]\label{thm:error-bound}
    With the choice of 
    \begin{gather}
        B_{K-M}\gets \frac{\gamma\sqrt{d}}{K-M-L+1} \left(\left(K-M-L\right)\ln \left(2n\right) + \ln\theta\right),\\
        \theta^{(k)} \gets\exp\left(\frac{B^{(k)}}{\gamma\sqrt{d}}(b^{(k)}-1)\right) \left(\theta\exp\left(-\frac{B_{K-M}-B_L}{\gamma\sqrt{d}}\right)\right)^\frac{1}{K-M-L}
    \end{gather} and irrelevant of the choice of other $B_k$s, the error bound in \eqref{eq:error-bound} can be minimized as \begin{equation}
        \varepsilon_\text{attn} = O\left(\left(K-M-L\right)\left(\frac{n}{\theta}\right)^\frac{1}{K-M-L+1}\right).
    \end{equation}
\end{Thm}

 Theorem \ref{thm:error-bound} have multiple implications: \begin{enumerate}
    \item Minimizing $K-M-L$, (i.e., the number of segments that are designated as neither the most significant nor the least significant as in Section \ref{sec:zkattn-optimization}) and $B_L$ (i.e., the magnitude of least significant segments) is in favour of reducing the error. However, as will be discussed in Section \ref{sec:oh-anal}, this incurs an undue increase in the computational overhead due to the sizes of the lookup tables.
    \item $\varepsilon_\text{attn}$ has no dependence on the segmentation of the \tt{zkAttn} input except $B_L$ and $B_{K-M}$, leaving room for distributing the sizes of lookup tables $\mathbf{T}^{(k)}$ evenly which compresses the computational overhead associated.
    \item $\varepsilon_\text{attn}$ defines the tolerable error row-wise sum upon checking the normalization, i.e., the sum of each row must lie within $[(1-\varepsilon_\text{attn})\theta, (1+\varepsilon_\text{attn})\theta]$.
\end{enumerate}

\subsection{Security and privacy analysis} \label{sec:sp-anal}

In this section, we formalize the security and privacy aspects of zkLLM, focusing on the \texttt{tlookup} protocol which facilitates zero-knowledge verifiable computations across all non-arithmetic components. We first address the completeness error of \texttt{tlookup}:

\begin{Thm}[Completeness of Protocol \ref{protocol:tlookup}] \label{thm:tlookup-completeness}
    Assuming the verifier \(\mathcal{V}\) is semi-honest, Protocol \ref{protocol:tlookup} incurs a completeness error of \(O\left(\frac{N}{|\mathbb{F}|}\right)\).  
\end{Thm}

Theorem \ref{thm:tlookup-completeness} indicates that if the prover \(\mathcal{P}\) follows the protocol, the proof produced using Protocol \ref{protocol:tlookup} has a mere \(O\left(\frac{N}{|\mathbb{F}|}\right)\) chance of being rejected. As detailed in Appendix \ref{appendix:sp-anal}, this minor imperfection stems from the probability that the random challenge \(\beta\), chosen by the verifier in Line \ref{alg-line:tlookup-random-challenge}, might trigger a division-by-zero error in one of the terms on either side of Equation \eqref{eq:hab22}. On the other hand, all arithmetic tensor operations do not contribute any additional completeness error, thanks to the direct application of the sumcheck protocol. Therefore, the overall probability of an honest prover failing to convince a semi-honest verifier of the correctness of its inference through the LLM is at most \(O\left(\frac{C}{|\mathbb{F}|}\right)\), where \(C\) represents the total size of all tensors involved in non-arithmetic operations. Given that the entire complexity of the inference is \(\text{poly}(\lambda)\) and \(|\F| = \Omega\left(2^{2\lambda}\right)\), this error remains negligible in \(\lambda\). Practically, as our implementation employs the BLS12-381 curve with \(|\F|\approx 2^{254}\), far exceeding the computational limits of current technology, verification has never failed in any experiment conducted, as reported in Section \ref{sec:experiments}.

Similarly, the soundness error of \texttt{tlookup} is also negligible in \(\lambda\), as stated in Theorem \ref{thm:tlookup-soundness}. Coupled with the direct application of the sumcheck protocol and the proof-of-opening for the committed tensors, which also incur negligible errors in \(\lambda\) due to the polynomial \(\lambda\) assumption on the complexity of the entire inference process, we theoretically establish that a valid proof can confirm the correctness of the inference result except with only a negligible probability.

\begin{Thm}[Soundness of Protocol \ref{protocol:tlookup}] \label{thm:tlookup-soundness}
    For any probabilistic polynomial-time (p.p.t.) prover \(\mathcal{P}\), if in Line \ref{alg-line:tlookup-commit}, the message \(\mathcal{P}\) sends to \(\mathcal{V}\) is \(\dbbracket{\mathbf{S}}\gets \texttt{Commit}(\mathbf{S})\) such that \(\mathbf{S}\not\subset \mathbf{T}\), then except with probability \(\negl{\lambda}\), the execution of Protocol \ref{protocol:tlookup} is unsuccessful, resulting in the semi-honest verifier \(\mathcal{V}\) rejecting the proof.
\end{Thm}

\begin{proof}[Proof sketch of Theorem \ref{thm:tlookup-soundness}]
    By the binding property of the commitment scheme, except with probability \(\negl{\lambda}\), in Line \ref{alg-line:tlookup-sumchecks}, the success of proofs of evaluations implies the correctness of all claimed multilinear extension values on \(\mathbf{A}, \mathbf{B}, \mathbf{S}, \mathbf{m}\), and \(\mathbf{T}\). Subsequently, the success of the sumchecks implies with \(1 - O\left(\frac{D}{|\mathbb{F}|}\right)\) that all equalities in Equations \eqref{eq:hab22-invs} and \eqref{eq:hab22-check} hold, such that 
    \begin{equation}
        \sum_{i \in \bracket{D}} \frac{1}{\beta + \mathbf{S}_i} = \sum_{i \in \bracket{N}} \frac{\mathbf{m}_i}{\beta + \mathbf{T}_i}. \label{eq:hab22-rand}
    \end{equation}
    Finally, given the randomness of \(\beta\), with probability \(1-O\left(\frac{ND}{|\mathbb{F}|}\right)\), Equation \eqref{eq:hab22-rand} implies Equation \eqref{eq:hab22}, leading to the conclusion that \(\mathbf{S} \subset \mathbf{T}\).
\end{proof}

It is noteworthy, as elaborated in Section \ref{sec:error-anal}, that the correctness of \texttt{zkAttn} is quantified by an \(\mathcal{L}_1\) error of \(\varepsilon_\text{attn}\). This measure of correctness similarly applies to other exponentiation-based activation functions. The correctness of all other non-arithmetic operations must also consider the quantization errors. For example, as highlighted in Example \ref{exm:relu}, a numerical error margin of \(\frac{1}{2\gamma}\) is an inescapable consequence of the rescaling process. On the other hand, this degree of tolerance is also sufficient, as numerical errors exceeding \(\frac{1}{2\gamma}\) would be detectable as incorrect computations and consequently rejected through the application of \texttt{tlookup}.

Finally, with the application of zero-knowledge variations of sumcheck protocols \cite{DBLP:journals/eccc/ChiesaFS17, libra, orion, zkcnn} and Pedersen commitment schemes \cite{pedersen}, the proof assembled by zkLLM does not disclose any information about the protected model parameters. Formally,

\begin{Thm}[Zero-knowledge, adapted from \cite{zkcnn}]
    Assuming the application of zero-knowledge variations of sumcheck protocols \cite{DBLP:journals/eccc/ChiesaFS17, libra, orion, zkcnn} and Pedersen commitment schemes \cite{pedersen}, there exists a simulator \(\mathcal{S} = (\mathcal{S}_1, \mathcal{S}_2)\) such that the following two views are computationally indistinguishable to any probabilistic polynomial-time (PPT) algorithm \(\mathcal{A}\), given the public parameters \(\texttt{pp}\) (the generators used in the commitment scheme within the context of zkLLM):

    \begin{mdframed}[backgroundcolor=white,linewidth=1pt,roundcorner=5pt]
    \noindent\(\texttt{Real}_{\mathcal{A}, \mathbf{W}}(\texttt{pp})\):
      \begin{algorithmic}[1]
        \State \(\com{\mathbf{W}} \gets \texttt{zkLLM-Commit}(\mathbf{W}, \texttt{pp}, r)\)
        \State \(\mathbf{X} \gets \mathcal{A}(\com{\mathbf{W}}, \texttt{pp})\)
        \State \((y, \pi) \gets \texttt{zkLLM-Prove}(\mathbf{W}, \mathbf{X}, \texttt{pp}, r)\)
        \State \(b \gets \mathcal{A}(\com{\mathbf{W}}, \mathbf{X}, y, \pi, \texttt{pp})\)
        \State \Return \(b\)
      \end{algorithmic}
    \end{mdframed}
    
    \begin{mdframed}[backgroundcolor=white,linewidth=1pt,roundcorner=5pt]
    \noindent\(\texttt{Ideal}_{\mathcal{A}, \mathcal{S}^\mathcal{A}}(\texttt{pp})\):
      \begin{algorithmic}[1]
        \State \(\texttt{com} \gets \mathcal{S}_1(1^\lambda, \texttt{pp}, r)\)
        \State \(\mathbf{X} \gets \mathcal{A}(\texttt{com}, \texttt{pp})\)
        \State \((y, \pi) \gets \mathcal{S}_2^\mathcal{A}(\texttt{com}, \mathbf{X}, \texttt{pp}, r)\), given oracle access to \(y = \texttt{zkLLM-compute}(\mathbf{W}, \mathbf{X})\)
        \State \(b \gets \mathcal{A}(\texttt{com}, \mathbf{X}, y, \pi, \texttt{pp})\)
        \State \Return \(b\)
      \end{algorithmic}
    \end{mdframed}
    
    For any PPT algorithm \(\mathcal{A}\) and all LLM (represented by the parameter) \(\mathbf{W}\), there exists a simulator \(\mathcal{S}\) such that
    \begin{equation}
        \abs{\mathbb{P}\left(\texttt{Real}_{\mathcal{A}, \mathbf{W}}(\texttt{pp}) = 1\right) - \mathbb{P}\left(\texttt{Ideal}_{\mathcal{A}, \mathcal{S}^\mathcal{A}}(\texttt{pp}) = 1\right)} \leq \negl{\lambda}.
    \end{equation}
\end{Thm}

\subsection{Overhead analysis} \label{sec:oh-anal}

In this section, we analyze the overhead of zkLLM, focusing on the running times for both the prover and the verifier, as well as the memory and communication costs.

\subsubsection{Overhead of \texttt{tlookup}} \label{sec:oh-tlookup}
In Protocol \ref{protocol:tlookup}, we adhere to the assumption that \(N = O(D)\). This guideline is strictly followed in our implementation due to the substantial sizes of tensors involved in LLM computations. The linear time complexity for both committing and proving in the Pedersen commitments and the sumcheck protocols results in a total computational complexity of \(O(D)\) for the prover in Protocol \ref{protocol:tlookup}. Similarly, memory requirements are maintained at \(O(D)\), since all involved tensors, including the additionally computed \(\mathbf{m}\), \(\mathbf{A}\), and \(\mathbf{B}\), are of size \(O(D)\). The commitment and proof sizes are reduced to square root and logarithmic complexities, \(O(\sqrt{D})\) and \(O(\log D)\) respectively, impacting the verifier's time for verifying the proof of opening and the sumcheck protocols.

\subsubsection{Overhead of \texttt{zkAttn}} \label{sec:oh-zkattn}
The overhead introduced by \texttt{zkAttn} is parameterized by \(K\), the number of segments applied to the input. For an input of size \(mn\) in \texttt{zkAttn}, the total prover overhead, including both running time and memory consumption, is \(O(Kmn)\) as per Section \ref{sec:oh-tlookup}. The communication overhead and verifier time are \(O(K\sqrt{mn})\). In comparison with the bit-decomposition method, which incurs an \(\Omega(mn\log_2 B)\) overhead, \texttt{zkAttn} built upon \texttt{tlookup} achieves an \(O\left(\frac{K}{\log_2B}\right)\) reduction in prover overhead. Practically, \(K\) is chosen to be small enough to minimize both error and overhead while avoiding overly large segments (for example, \(K=1\), which would require compiling all possible input-output pairs into a table, leading to an impractical overhead of at least \(\Omega(B)\) for a unrealistically large total number of possible inputs \(B\) that exceeds \(mn\), where the previous analysis on overhead would not apply).

\subsubsection{Overall overhead} \label{sec:oh-overall}
zkLLM benefits from the linear prover overhead of sumcheck protocols and the logarithmic and square-root verifier overheads of sumchecks and Pedersen commitments, respectively. Specialized sumchecks for tensor operations, like matrix multiplications, achieve less complexity than the computation process itself, further reducing proof overhead for each layer. Assuming prover overhead, communication cost, and verifier overhead per layer of an \(L\)-layer LLM are \(t_\mathcal{P}\), \(c\), and \(t_\mathcal{V}\) respectively, the total overheads of zkLLM scale naturally to \(O(L t_\mathcal{P})\), \(O(Lc)\), and \(O(Lt_\mathcal{V})\). Where the latter two can be further reduced to \(O(\sqrt{L}C)\) and \(O(\sqrt{L}t_\mathcal{V})\) by leveraging the repetitive structure across layers and batching up the commitments \cite{zkcnn}. Additionally, unlike univariate polynomial-based ZKP systems that must be serialized, the use of sumcheck protocols over multilinear extensions and the compatible Pedersen commitment scheme in zkLLM allows for highly parallelized proof generation, thereby enabling efficient proof generation in a reasonable time.

\section{Experiments} \label{sec:experiments}

\begin{table*}[htbp]
    \centering
    \caption{The overhead of zkLLM on OPT and LLaMa-2.}
    \begin{tabular}{lcccccccc}
        Model & OPT-125M & OPT-350M & OPT-1.3B & OPT-2.7B & OPT-6.7B & OPT-13B & LLaMa-2-7B & LLaMa-2-13B \\
        \hline
        Committing time (s) & 11.8 & 33.1 & 127 & 273 & 654 & $1.27\times 10^3$ & 531 & 986 \\
        Commitment size (MB) & 0.996 & 1.67 & 3.32 & 4.58 & 7.22 & 10.1 & 7.97 & 11.0 \\
        Prover time (s) & 73.9 & 111 & 221 & 352 & 548 & 713 & 620 & 803\\
        Proof size (kB) & 141 & 144 & 147 & 152 & 157 & 160 & 183 & 188 \\
        Verifier time (s) & 0.342 & 0.593 & 0.899 & 1.41 & 2.08 & 3.71 & 2.36 & 3.95 \\
        Memory usage (GB) & 1.88 & 2.38 & 3.71 & 6.60 & 15.0 & 22.9 & 15.5 & 23.1 \\
        C4 Perplexity (orig) & 26.56 & 22.59 & 16.07 & 14.34 & 12.71 & 12.06 & 7.036 & 6.520 \\
        C4 Perplexity (quant) & 26.65 & 22.66 & 16.12 & 14.37 & 12.73 & 12.07 & 7.049 & 6.528 
    \end{tabular}
    \label{tab:results}
\end{table*}

We developed zkLLM using CUDA, basing it on the CUDA code for the BLS12-381 curve \cite{bls} produced by the \texttt{ec-gpu} package \cite{ec-gpu}. The implementation of sequential verifier tasks, which cannot efficiently utilize CUDA, was adapted from the zkCNN implementation \cite{zkcnn} that relies on the \texttt{mcl} package \cite{mcl}. We evaluated zkLLM for inferences on two classes of open-source LLMs, namely OPT \cite{opt} and LLaMa-2 \cite{llama2}, supporting sizes up to 13 billion parameters. For both types of models, our focus was on performing verifiable inferences using the designated models, applying samples with the default sequence length of 2048 from the C4 dataset \cite{c4}. Our experiments were conducted with resources including 124.5GB of memory, 12 CPU cores of an AMD EPYC 7413 (2.65 GHz with 128M cache L3), and an NVIDIA A100SMX4 GPU with 40GB of memory, all allocated from a computing node.

Throughout our experiments, we consistently set the scaling factor for both data embedding and model parameters at \(2^{16}\). All rescaling operations were integrated with the subsequent activation functions. As detailed in Section \ref{exm:relu}, this integration necessitated the use of multiple \texttt{tlookup}s. Specifically, the number of \texttt{tlookup}s corresponds to the number of times the input to the activation function requires rescaling, with each \texttt{tlookup} having a size of \(2^{16}\), to avoid excessive memory usage. 

Additionally, since the input to the Softmax function in \texttt{zkAttn} of each layer undergoes two multiplication operations, the cumulative scaling factor reaches \(2^{64}\). To manage this, we deployed \(K = 5\) \texttt{tlookup}s, each of size \(2^{16}\). This setup includes \(L = 3\) least significant segments, with the remaining two segments accommodating all potential inputs within a scale of \(2^{16}\) and a precision of \(2^{-16}\) when reverted to the real domain. 

The tolerable error margins were selected in accordance with Section \ref{sec:error-anal}. This approach resulted in an approximate total \(\mathcal{L}_1\) error of \(10^{-2}\) on the output, a level comparable to the rounding error induced by half-precision floating points used in state-of-the-art LLMs. Notably, all proofs involved in the results of this section have been successfully validated by the semi-honest verifier.

In Table \ref{tab:results}, we present detailed information regarding the overhead associated with zkLLM for models of various sizes. The data includes the time required for the prover to commit and hide the model (committing time and commitment sizes), as well as the prover's time, proof size, and the verifier's time in response to an input prompt from the latter.

As the first endeavor to apply zero-knowledge proofs to LLMs with up to 13 billion parameters, to the best of our knowledge, zkLLM has achieved significant results toward practical and industrial applicability. Once the model is trained, the prover requires up to 20 minutes to commit to the model and subsequently publish a commitment of approximately 10MB for public scrutiny via zkLLM. When prompted by the verifier, the prover is able to generate a proof for the correctness of the entire inference process in less than 15 minutes. Additionally, as the design of \texttt{zkAttn} effectively resolves the inherent bottleneck of listing all input-output pairs, the memory consumption is effectively controlled under 23.1GB, which fits zkLLM into commonly used GPUs in machine learning, like Tesla V100 and A100.

It is important to note that while the time for committing generally scales with the size of the parameters, the time for generating proofs scales more slowly. This slower scaling is attributed to the less significant differences in the complexities of intermediate computations and more efficient use of parallel computing resources as the size of the tensors increases. Ultimately, the succinct proof, which is only about 100kB in size, can be verified by the verifier in a matter of seconds. Despite this efficiency, the verifier is provably unable to glean any additional information about the model or the inference result, ensuring the correctness of the inference while maintaining the confidentiality of the model parameters.

Moreover, the numerical error due to the unavoidable discretization of the entire process for the application of the cryptographic tools does not cause significant accuracy drops: on the C4 dataset \cite{c4}, the increase of perplexity is less than 0.1, and the impact diminishes to less than 0.01 as the sizes of the models scale to 13B.

\begin{figure}
    \centering
    \includegraphics[width=\linewidth]{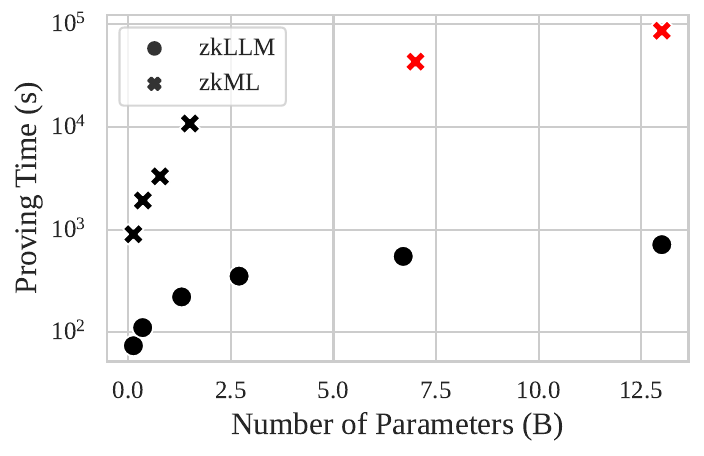}
    \caption{Comparison between zkLLM and zkML. Results of test cases with OOM errors are estimated and marked in red.}
    \label{fig:compare-zkml}
\end{figure}

In Figure \ref{fig:compare-zkml}, we further compare zkLLM with zkML \cite{zkml}, the first zero-knowledge proof to have achieved verifiable inference for GPT-2 under identical hardware conditions. Beyond the size of GPT-2 (1.5B parameters), where zkML results in an out-of-memory (OOM) error, we provide an estimation of the required proving time. Thanks to the design tailored for non-arithmetic tensor operations and the attention mechanism prevalent in LLMs, as well as its CUDA implementation, zkLLM extends the zero-knowledge verifiability to LLMs with 10x larger sizes, achieving an approximate 50x speedup.

\subsection{Additional experimental results on \texttt{tlookup} and \texttt{zkAttn}}
\begin{figure*}[htbp]
    \centering
    \includegraphics[width=\linewidth]{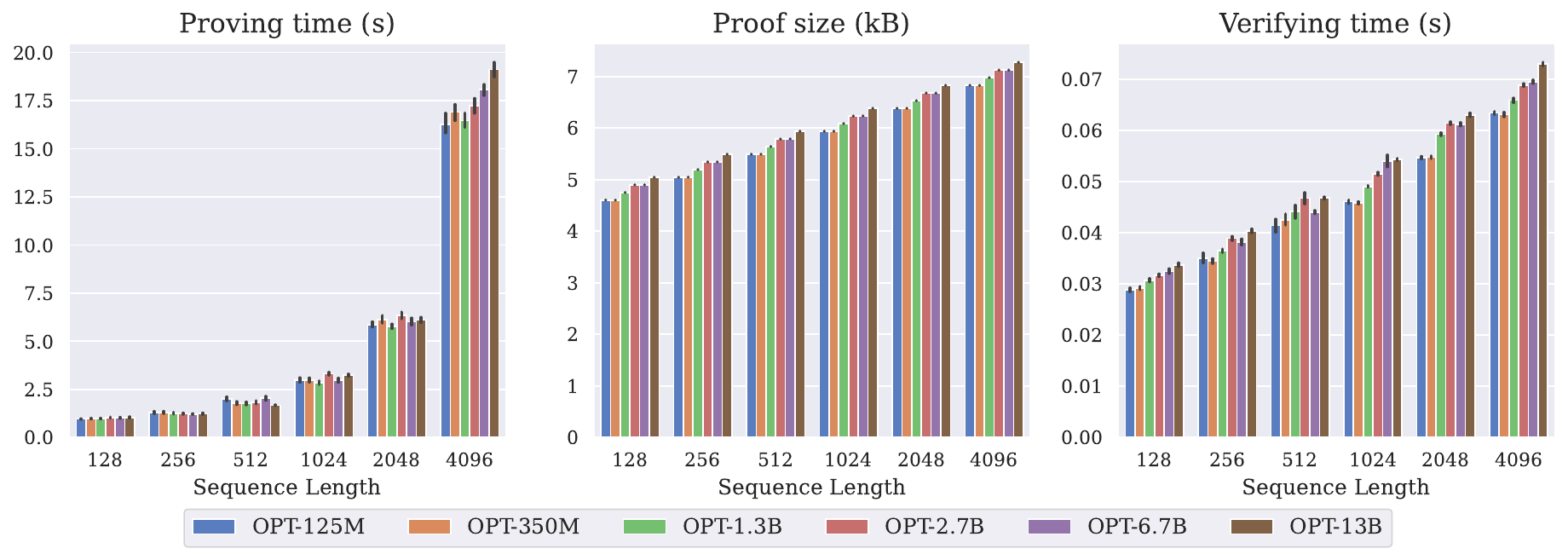}
    \caption{Overhead of \texttt{zkAttn}.}
    \label{fig:zkattn-overhead}
\end{figure*}

To further demonstrate the efficiency of \texttt{tlookup} in addressing non-arithmetic tensor operations in deep learning, as well as \texttt{zkAttn}, which is pivotal for verifiable computations in LLMs, we isolated an instance of \texttt{zkAttn} from the first layer of variously sized OPT models on input sequences of different lengths. We then measured the overhead incurred by \texttt{zkAttn}, including the multiple \texttt{tlookup}s it encompasses. The results are presented in Figure \ref{fig:zkattn-overhead}.

It is evident that, in contrast to the overall overhead, the overhead specific to \texttt{zkAttn} is less influenced by the size of the model, particularly regarding proving time. However, the length of the input sequence significantly impacts various aspects of the overhead. This observation can be attributed to the fact that the Attention mechanism, while involving LLM parameters, is more significantly affected by the interactions between intermediate values (e.g., \(\mathbf{Q}, \mathbf{K}, \mathbf{V}\)), where their dimensions play a crucial role in determining the overhead.

Notably, the largest tested sequence length, 4096, exceeds the original design specifications of OPT models and was primarily included as a reference to assess the impact of sequence length on overhead. In contrast, in Table \ref{tab:results}, which documents overall results for zkLLM, we set the sequence length to 2048—the maximum feasible value—to maintain experimental consistency and fairness.

\section{Related works} \label{sec:related-works}

Starting with Zhang et al. in 2020 \cite{DBLP:conf/ccs/ZhangFZS20}, the field of zero-knowledge machine learning inference has seen active development. Initial research, parallel to the surge in computer vision studies, primarily concentrated on authenticating inference results for computer vision tasks over convolutional neural networks (CNNs). Key contributions include zkCNN \cite{zkcnn}, ZEN \cite{zen}, vCNN \cite{vcnn}, pvCNN \cite{pvcnn}, zkML \cite{zkml}, Mystique \cite{mystique}, and ezDPS \cite{ezdps}. These works aimed to optimize the adaptation of the entire training process to zero-knowledge proof (ZKP) backends, such as zkSNARKS \cite{pinocchio,groth16,plonk,halo,halo-inf,DBLP:journals/joc/BitanskyCIOP22,hyperplonk,ligero,ligero++}, by leveraging the special structures of computations within CNNs. Notably, zkCNN \cite{zkcnn} introduced a specialized interactive proof protocol for convolutional layers based on the GKR protocol \cite{gkr} and its refinements \cite{libra,DBLP:conf/ccs/ZhangLWZSXZ21,orion}. This protocol achieved efficient proofs (less than 2 minutes) on VGG-scale CNNs, highlighting the necessity of specialized protocols for realistic zero-knowledge machine learning inference. However, as of our current knowledge, there exists a gap in zero-knowledge inferences over LLMs. The intricate structures and enormous sizes of LLMs present challenges not addressed by previous studies focused on CNNs, necessitating novel theoretical and experimental developments.

Conversely, while pioneering studies have addressed ZKPs for machine learning training, with works such as VeriML \cite{veriml}, proof of unlearning \cite{unlearning,poul}, and zkPoT \cite{zkpot} focusing on elementary algorithms like Support Vector Machines (SVM), logistic regression, and small neural networks (up to several thousand parameters), extending zero-knowledge proofs to training LLMs may pose insurmountable challenges. The vast complexity inherent in training LLMs could render the zero-knowledge proofs for their training impractical.

\section{Conclusion} \label{sec:conclusion}

This paper introduces zkLLM, marking the inaugural specialized zero-knowledge proof tailored for large language models, as far as our current knowledge extends. zkLLM pioneers zero-knowledge verifiable computations for general non-arithmetic operations within neural networks, ensuring full parallelizability and zero additional overhead through the implementation of \tt{tlookup}. Building on this foundation, we further present \tt{zkAttn}, a novel zero-knowledge proof specifically designed for the attention mechanism—a pivotal component underpinning the exceptional performance of modern LLMs. With a CUDA implementation optimized for parallel computing resources in deep learning, zkLLM achieves a groundbreaking milestone as the first study to provide zero-knowledge verifiability for LLMs with 13 billion parameters. This endeavor stands as a significant contribution towards fortifying the legitimacy of LLMs in light of their transformative impact on various domains.

\bibliographystyle{ACM-Reference-Format}

\bibliography{reference}


\begin{thebibliography}{69}


\ifx \showCODEN    \undefined \def \showCODEN     #1{\unskip}     \fi
\ifx \showDOI      \undefined \def \showDOI       #1{#1}\fi
\ifx \showISBNx    \undefined \def \showISBNx     #1{\unskip}     \fi
\ifx \showISBNxiii \undefined \def \showISBNxiii  #1{\unskip}     \fi
\ifx \showISSN     \undefined \def \showISSN      #1{\unskip}     \fi
\ifx \showLCCN     \undefined \def \showLCCN      #1{\unskip}     \fi
\ifx \shownote     \undefined \def \shownote      #1{#1}          \fi
\ifx \showarticletitle \undefined \def \showarticletitle #1{#1}   \fi
\ifx \showURL      \undefined \def \showURL       {\relax}        \fi
\providecommand\bibfield[2]{#2}
\providecommand\bibinfo[2]{#2}
\providecommand\natexlab[1]{#1}
\providecommand\showeprint[2][]{arXiv:#2}

\bibitem[Anil et~al\mbox{.}(2023)]%
        {gemini}
\bibfield{author}{\bibinfo{person}{Rohan Anil}, \bibinfo{person}{Sebastian Borgeaud}, \bibinfo{person}{Yonghui Wu}, \bibinfo{person}{Jean{-}Baptiste Alayrac}, \bibinfo{person}{Jiahui Yu}, \bibinfo{person}{Radu Soricut}, \bibinfo{person}{Johan Schalkwyk}, \bibinfo{person}{Andrew~M. Dai}, \bibinfo{person}{Anja Hauth}, \bibinfo{person}{Katie Millican}, \bibinfo{person}{David Silver}, \bibinfo{person}{Slav Petrov}, \bibinfo{person}{Melvin Johnson}, \bibinfo{person}{Ioannis Antonoglou}, \bibinfo{person}{Julian Schrittwieser}, \bibinfo{person}{Amelia Glaese}, \bibinfo{person}{Jilin Chen}, \bibinfo{person}{Emily Pitler}, \bibinfo{person}{Timothy~P. Lillicrap}, \bibinfo{person}{Angeliki Lazaridou}, \bibinfo{person}{Orhan Firat}, \bibinfo{person}{James Molloy}, \bibinfo{person}{Michael Isard}, \bibinfo{person}{Paul~Ronald Barham}, \bibinfo{person}{Tom Hennigan}, \bibinfo{person}{Benjamin Lee}, \bibinfo{person}{Fabio Viola}, \bibinfo{person}{Malcolm Reynolds}, \bibinfo{person}{Yuanzhong Xu}, \bibinfo{person}{Ryan
  Doherty}, \bibinfo{person}{Eli Collins}, \bibinfo{person}{Clemens Meyer}, \bibinfo{person}{Eliza Rutherford}, \bibinfo{person}{Erica Moreira}, \bibinfo{person}{Kareem Ayoub}, \bibinfo{person}{Megha Goel}, \bibinfo{person}{George Tucker}, \bibinfo{person}{Enrique Piqueras}, \bibinfo{person}{Maxim Krikun}, \bibinfo{person}{Iain Barr}, \bibinfo{person}{Nikolay Savinov}, \bibinfo{person}{Ivo Danihelka}, \bibinfo{person}{Becca Roelofs}, \bibinfo{person}{Ana{\"{\i}}s White}, \bibinfo{person}{Anders Andreassen}, \bibinfo{person}{Tamara von Glehn}, \bibinfo{person}{Lakshman Yagati}, \bibinfo{person}{Mehran Kazemi}, \bibinfo{person}{Lucas Gonzalez}, \bibinfo{person}{Misha Khalman}, \bibinfo{person}{Jakub Sygnowski}, {and} \bibinfo{person}{et al.}} \bibinfo{year}{2023}\natexlab{}.
\newblock \showarticletitle{Gemini: {A} Family of Highly Capable Multimodal Models}.
\newblock \bibinfo{journal}{\emph{CoRR}}  \bibinfo{volume}{abs/2312.11805} (\bibinfo{year}{2023}).
\newblock
\urldef\tempurl%
\url{https://doi.org/10.48550/ARXIV.2312.11805}
\showDOI{\tempurl}
\showeprint[arXiv]{2312.11805}


\bibitem[Ba et~al\mbox{.}(2016)]%
        {layernorm}
\bibfield{author}{\bibinfo{person}{Lei~Jimmy Ba}, \bibinfo{person}{Jamie~Ryan Kiros}, {and} \bibinfo{person}{Geoffrey~E. Hinton}.} \bibinfo{year}{2016}\natexlab{}.
\newblock \showarticletitle{Layer Normalization}.
\newblock \bibinfo{journal}{\emph{CoRR}}  \bibinfo{volume}{abs/1607.06450} (\bibinfo{year}{2016}).
\newblock
\showeprint[arXiv]{1607.06450}
\urldef\tempurl%
\url{http://arxiv.org/abs/1607.06450}
\showURL{%
\tempurl}


\bibitem[Bhadauria et~al\mbox{.}(2020)]%
        {ligero++}
\bibfield{author}{\bibinfo{person}{Rishabh Bhadauria}, \bibinfo{person}{Zhiyong Fang}, \bibinfo{person}{Carmit Hazay}, \bibinfo{person}{Muthuramakrishnan Venkitasubramaniam}, \bibinfo{person}{Tiancheng Xie}, {and} \bibinfo{person}{Yupeng Zhang}.} \bibinfo{year}{2020}\natexlab{}.
\newblock \showarticletitle{Ligero++: {A} New Optimized Sublinear {IOP}}. In \bibinfo{booktitle}{\emph{{CCS} '20: 2020 {ACM} {SIGSAC} Conference on Computer and Communications Security, Virtual Event, USA, November 9-13, 2020}}, \bibfield{editor}{\bibinfo{person}{Jay Ligatti}, \bibinfo{person}{Xinming Ou}, \bibinfo{person}{Jonathan Katz}, {and} \bibinfo{person}{Giovanni Vigna}} (Eds.). \bibinfo{publisher}{{ACM}}, \bibinfo{pages}{2025--2038}.
\newblock
\urldef\tempurl%
\url{https://doi.org/10.1145/3372297.3417893}
\showDOI{\tempurl}


\bibitem[Bitansky et~al\mbox{.}(2022)]%
        {DBLP:journals/joc/BitanskyCIOP22}
\bibfield{author}{\bibinfo{person}{Nir Bitansky}, \bibinfo{person}{Alessandro Chiesa}, \bibinfo{person}{Yuval Ishai}, \bibinfo{person}{Rafail Ostrovsky}, {and} \bibinfo{person}{Omer Paneth}.} \bibinfo{year}{2022}\natexlab{}.
\newblock \showarticletitle{Succinct Non-Interactive Arguments via Linear Interactive Proofs}.
\newblock \bibinfo{journal}{\emph{J. Cryptol.}} \bibinfo{volume}{35}, \bibinfo{number}{3} (\bibinfo{year}{2022}), \bibinfo{pages}{15}.
\newblock
\urldef\tempurl%
\url{https://doi.org/10.1007/Ss00145-022-09424-4}
\showDOI{\tempurl}


\bibitem[Boneh et~al\mbox{.}(2020)]%
        {halo-inf}
\bibfield{author}{\bibinfo{person}{Dan Boneh}, \bibinfo{person}{Justin Drake}, \bibinfo{person}{Ben Fisch}, {and} \bibinfo{person}{Ariel Gabizon}.} \bibinfo{year}{2020}\natexlab{}.
\newblock \showarticletitle{Halo Infinite: Recursive zk-{SNARKs} from any Additive Polynomial Commitment Scheme}.
\newblock \bibinfo{journal}{\emph{{IACR} Cryptol. ePrint Arch.}} (\bibinfo{year}{2020}), \bibinfo{pages}{1536}.
\newblock
\urldef\tempurl%
\url{https://eprint.iacr.org/2020/1536}
\showURL{%
\tempurl}


\bibitem[Boneh et~al\mbox{.}(2001)]%
        {bls}
\bibfield{author}{\bibinfo{person}{Dan Boneh}, \bibinfo{person}{Ben Lynn}, {and} \bibinfo{person}{Hovav Shacham}.} \bibinfo{year}{2001}\natexlab{}.
\newblock \showarticletitle{Short Signatures from the Weil Pairing}. In \bibinfo{booktitle}{\emph{Advances in Cryptology - {ASIACRYPT} 2001, 7th International Conference on the Theory and Application of Cryptology and Information Security, Gold Coast, Australia, December 9-13, 2001, Proceedings}} \emph{(\bibinfo{series}{Lecture Notes in Computer Science}, Vol.~\bibinfo{volume}{2248})}, \bibfield{editor}{\bibinfo{person}{Colin Boyd}} (Ed.). \bibinfo{publisher}{Springer}, \bibinfo{pages}{514--532}.
\newblock
\urldef\tempurl%
\url{https://doi.org/10.1007/3-540-45682-1\_30}
\showDOI{\tempurl}


\bibitem[Bowe et~al\mbox{.}(2019)]%
        {halo}
\bibfield{author}{\bibinfo{person}{Sean Bowe}, \bibinfo{person}{Jack Grigg}, {and} \bibinfo{person}{Daira Hopwood}.} \bibinfo{year}{2019}\natexlab{}.
\newblock \showarticletitle{Halo: Recursive Proof Composition without a Trusted Setup}.
\newblock \bibinfo{journal}{\emph{{IACR} Cryptol. ePrint Arch.}} (\bibinfo{year}{2019}), \bibinfo{pages}{1021}.
\newblock
\urldef\tempurl%
\url{https://eprint.iacr.org/2019/1021}
\showURL{%
\tempurl}


\bibitem[Brown et~al\mbox{.}(2020)]%
        {gpt3}
\bibfield{author}{\bibinfo{person}{Tom~B. Brown}, \bibinfo{person}{Benjamin Mann}, \bibinfo{person}{Nick Ryder}, \bibinfo{person}{Melanie Subbiah}, \bibinfo{person}{Jared Kaplan}, \bibinfo{person}{Prafulla Dhariwal}, \bibinfo{person}{Arvind Neelakantan}, \bibinfo{person}{Pranav Shyam}, \bibinfo{person}{Girish Sastry}, \bibinfo{person}{Amanda Askell}, \bibinfo{person}{Sandhini Agarwal}, \bibinfo{person}{Ariel Herbert{-}Voss}, \bibinfo{person}{Gretchen Krueger}, \bibinfo{person}{Tom Henighan}, \bibinfo{person}{Rewon Child}, \bibinfo{person}{Aditya Ramesh}, \bibinfo{person}{Daniel~M. Ziegler}, \bibinfo{person}{Jeffrey Wu}, \bibinfo{person}{Clemens Winter}, \bibinfo{person}{Christopher Hesse}, \bibinfo{person}{Mark Chen}, \bibinfo{person}{Eric Sigler}, \bibinfo{person}{Mateusz Litwin}, \bibinfo{person}{Scott Gray}, \bibinfo{person}{Benjamin Chess}, \bibinfo{person}{Jack Clark}, \bibinfo{person}{Christopher Berner}, \bibinfo{person}{Sam McCandlish}, \bibinfo{person}{Alec Radford}, \bibinfo{person}{Ilya Sutskever},
  {and} \bibinfo{person}{Dario Amodei}.} \bibinfo{year}{2020}\natexlab{}.
\newblock \showarticletitle{Language Models are Few-Shot Learners}. In \bibinfo{booktitle}{\emph{Advances in Neural Information Processing Systems 33: Annual Conference on Neural Information Processing Systems 2020, NeurIPS 2020, December 6-12, 2020, virtual}}, \bibfield{editor}{\bibinfo{person}{Hugo Larochelle}, \bibinfo{person}{Marc'Aurelio Ranzato}, \bibinfo{person}{Raia Hadsell}, \bibinfo{person}{Maria{-}Florina Balcan}, {and} \bibinfo{person}{Hsuan{-}Tien Lin}} (Eds.).
\newblock
\urldef\tempurl%
\url{https://proceedings.neurips.cc/paper/2020/hash/1457c0d6bfcb4967418bfb8ac142f64a-Abstract.html}
\showURL{%
\tempurl}


\bibitem[Chee et~al\mbox{.}(2023)]%
        {quip}
\bibfield{author}{\bibinfo{person}{Jerry Chee}, \bibinfo{person}{Yaohui Cai}, \bibinfo{person}{Volodymyr Kuleshov}, {and} \bibinfo{person}{Christopher~De Sa}.} \bibinfo{year}{2023}\natexlab{}.
\newblock \showarticletitle{QuIP: 2-Bit Quantization of Large Language Models With Guarantees}.
\newblock \bibinfo{journal}{\emph{CoRR}}  \bibinfo{volume}{abs/2307.13304} (\bibinfo{year}{2023}).
\newblock
\urldef\tempurl%
\url{https://doi.org/10.48550/ARXIV.2307.13304}
\showDOI{\tempurl}
\showeprint[arXiv]{2307.13304}


\bibitem[Chen et~al\mbox{.}(2023)]%
        {hyperplonk}
\bibfield{author}{\bibinfo{person}{Binyi Chen}, \bibinfo{person}{Benedikt B{\"{u}}nz}, \bibinfo{person}{Dan Boneh}, {and} \bibinfo{person}{Zhenfei Zhang}.} \bibinfo{year}{2023}\natexlab{}.
\newblock \showarticletitle{HyperPlonk: Plonk with Linear-Time Prover and High-Degree Custom Gates}. In \bibinfo{booktitle}{\emph{Advances in Cryptology - {EUROCRYPT} 2023 - 42nd Annual International Conference on the Theory and Applications of Cryptographic Techniques, Lyon, France, April 23-27, 2023, Proceedings, Part {II}}} \emph{(\bibinfo{series}{Lecture Notes in Computer Science}, Vol.~\bibinfo{volume}{14005})}, \bibfield{editor}{\bibinfo{person}{Carmit Hazay} {and} \bibinfo{person}{Martijn Stam}} (Eds.). \bibinfo{publisher}{Springer}, \bibinfo{pages}{499--530}.
\newblock
\urldef\tempurl%
\url{https://doi.org/10.1007/978-3-031-30617-4\_17}
\showDOI{\tempurl}


\bibitem[Chiesa et~al\mbox{.}(2017)]%
        {DBLP:journals/eccc/ChiesaFS17}
\bibfield{author}{\bibinfo{person}{Alessandro Chiesa}, \bibinfo{person}{Michael~A. Forbes}, {and} \bibinfo{person}{Nicholas Spooner}.} \bibinfo{year}{2017}\natexlab{}.
\newblock \showarticletitle{A Zero Knowledge Sumcheck and its Applications}.
\newblock \bibinfo{journal}{\emph{Electron. Colloquium Comput. Complex.}}  \bibinfo{volume}{{TR17-057}} (\bibinfo{year}{2017}).
\newblock
\showeprint[ECCC]{TR17-057}
\urldef\tempurl%
\url{https://eccc.weizmann.ac.il/report/2017/057}
\showURL{%
\tempurl}


\bibitem[Chowdhery et~al\mbox{.}(2023)]%
        {palm}
\bibfield{author}{\bibinfo{person}{Aakanksha Chowdhery}, \bibinfo{person}{Sharan Narang}, \bibinfo{person}{Jacob Devlin}, \bibinfo{person}{Maarten Bosma}, \bibinfo{person}{Gaurav Mishra}, \bibinfo{person}{Adam Roberts}, \bibinfo{person}{Paul Barham}, \bibinfo{person}{Hyung~Won Chung}, \bibinfo{person}{Charles Sutton}, \bibinfo{person}{Sebastian Gehrmann}, \bibinfo{person}{Parker Schuh}, \bibinfo{person}{Kensen Shi}, \bibinfo{person}{Sasha Tsvyashchenko}, \bibinfo{person}{Joshua Maynez}, \bibinfo{person}{Abhishek Rao}, \bibinfo{person}{Parker Barnes}, \bibinfo{person}{Yi Tay}, \bibinfo{person}{Noam Shazeer}, \bibinfo{person}{Vinodkumar Prabhakaran}, \bibinfo{person}{Emily Reif}, \bibinfo{person}{Nan Du}, \bibinfo{person}{Ben Hutchinson}, \bibinfo{person}{Reiner Pope}, \bibinfo{person}{James Bradbury}, \bibinfo{person}{Jacob Austin}, \bibinfo{person}{Michael Isard}, \bibinfo{person}{Guy Gur{-}Ari}, \bibinfo{person}{Pengcheng Yin}, \bibinfo{person}{Toju Duke}, \bibinfo{person}{Anselm Levskaya},
  \bibinfo{person}{Sanjay Ghemawat}, \bibinfo{person}{Sunipa Dev}, \bibinfo{person}{Henryk Michalewski}, \bibinfo{person}{Xavier Garcia}, \bibinfo{person}{Vedant Misra}, \bibinfo{person}{Kevin Robinson}, \bibinfo{person}{Liam Fedus}, \bibinfo{person}{Denny Zhou}, \bibinfo{person}{Daphne Ippolito}, \bibinfo{person}{David Luan}, \bibinfo{person}{Hyeontaek Lim}, \bibinfo{person}{Barret Zoph}, \bibinfo{person}{Alexander Spiridonov}, \bibinfo{person}{Ryan Sepassi}, \bibinfo{person}{David Dohan}, \bibinfo{person}{Shivani Agrawal}, \bibinfo{person}{Mark Omernick}, \bibinfo{person}{Andrew~M. Dai}, \bibinfo{person}{Thanumalayan~Sankaranarayana Pillai}, \bibinfo{person}{Marie Pellat}, \bibinfo{person}{Aitor Lewkowycz}, \bibinfo{person}{Erica Moreira}, \bibinfo{person}{Rewon Child}, \bibinfo{person}{Oleksandr Polozov}, \bibinfo{person}{Katherine Lee}, \bibinfo{person}{Zongwei Zhou}, \bibinfo{person}{Xuezhi Wang}, \bibinfo{person}{Brennan Saeta}, \bibinfo{person}{Mark Diaz}, \bibinfo{person}{Orhan Firat},
  \bibinfo{person}{Michele Catasta}, \bibinfo{person}{Jason Wei}, \bibinfo{person}{Kathy Meier{-}Hellstern}, \bibinfo{person}{Douglas Eck}, \bibinfo{person}{Jeff Dean}, \bibinfo{person}{Slav Petrov}, {and} \bibinfo{person}{Noah Fiedel}.} \bibinfo{year}{2023}\natexlab{}.
\newblock \showarticletitle{PaLM: Scaling Language Modeling with Pathways}.
\newblock \bibinfo{journal}{\emph{J. Mach. Learn. Res.}}  \bibinfo{volume}{24} (\bibinfo{year}{2023}), \bibinfo{pages}{240:1--240:113}.
\newblock
\urldef\tempurl%
\url{http://jmlr.org/papers/v24/22-1144.html}
\showURL{%
\tempurl}


\bibitem[Cormode et~al\mbox{.}(2012)]%
        {sumcheck}
\bibfield{author}{\bibinfo{person}{Graham Cormode}, \bibinfo{person}{Michael Mitzenmacher}, {and} \bibinfo{person}{Justin Thaler}.} \bibinfo{year}{2012}\natexlab{}.
\newblock \showarticletitle{Practical verified computation with streaming interactive proofs}. In \bibinfo{booktitle}{\emph{Innovations in Theoretical Computer Science 2012, Cambridge, MA, USA, January 8-10, 2012}}, \bibfield{editor}{\bibinfo{person}{Shafi Goldwasser}} (Ed.). \bibinfo{publisher}{{ACM}}, \bibinfo{pages}{90--112}.
\newblock
\urldef\tempurl%
\url{https://doi.org/10.1145/2090236.2090245}
\showDOI{\tempurl}


\bibitem[Dettmers et~al\mbox{.}(2022)]%
        {llmint8}
\bibfield{author}{\bibinfo{person}{Tim Dettmers}, \bibinfo{person}{Mike Lewis}, \bibinfo{person}{Younes Belkada}, {and} \bibinfo{person}{Luke Zettlemoyer}.} \bibinfo{year}{2022}\natexlab{}.
\newblock \showarticletitle{LLM.int8(): 8-bit Matrix Multiplication for Transformers at Scale}.
\newblock \bibinfo{journal}{\emph{CoRR}}  \bibinfo{volume}{abs/2208.07339} (\bibinfo{year}{2022}).
\newblock
\urldef\tempurl%
\url{https://doi.org/10.48550/ARXIV.2208.07339}
\showDOI{\tempurl}
\showeprint[arXiv]{2208.07339}


\bibitem[Dettmers et~al\mbox{.}(2023)]%
        {qlora}
\bibfield{author}{\bibinfo{person}{Tim Dettmers}, \bibinfo{person}{Artidoro Pagnoni}, \bibinfo{person}{Ari Holtzman}, {and} \bibinfo{person}{Luke Zettlemoyer}.} \bibinfo{year}{2023}\natexlab{}.
\newblock \showarticletitle{QLoRA: Efficient Finetuning of Quantized LLMs}.
\newblock \bibinfo{journal}{\emph{CoRR}}  \bibinfo{volume}{abs/2305.14314} (\bibinfo{year}{2023}).
\newblock
\urldef\tempurl%
\url{https://doi.org/10.48550/ARXIV.2305.14314}
\showDOI{\tempurl}
\showeprint[arXiv]{2305.14314}


\bibitem[Eagen et~al\mbox{.}(2022)]%
        {cq}
\bibfield{author}{\bibinfo{person}{Liam Eagen}, \bibinfo{person}{Dario Fiore}, {and} \bibinfo{person}{Ariel Gabizon}.} \bibinfo{year}{2022}\natexlab{}.
\newblock \showarticletitle{cq: Cached quotients for fast lookups}.
\newblock \bibinfo{journal}{\emph{{IACR} Cryptol. ePrint Arch.}} (\bibinfo{year}{2022}), \bibinfo{pages}{1763}.
\newblock
\urldef\tempurl%
\url{https://eprint.iacr.org/2022/1763}
\showURL{%
\tempurl}


\bibitem[Eisenhofer et~al\mbox{.}(2022)]%
        {unlearning}
\bibfield{author}{\bibinfo{person}{Thorsten Eisenhofer}, \bibinfo{person}{Doreen Riepel}, \bibinfo{person}{Varun Chandrasekaran}, \bibinfo{person}{Esha Ghosh}, \bibinfo{person}{Olga Ohrimenko}, {and} \bibinfo{person}{Nicolas Papernot}.} \bibinfo{year}{2022}\natexlab{}.
\newblock \showarticletitle{Verifiable and Provably Secure Machine Unlearning}.
\newblock \bibinfo{journal}{\emph{CoRR}}  \bibinfo{volume}{abs/2210.09126} (\bibinfo{year}{2022}).
\newblock
\urldef\tempurl%
\url{https://doi.org/10.48550/arXiv.2210.09126}
\showDOI{\tempurl}
\showeprint[arXiv]{2210.09126}


\bibitem[Feng et~al\mbox{.}(2021)]%
        {zen}
\bibfield{author}{\bibinfo{person}{Boyuan Feng}, \bibinfo{person}{Lianke Qin}, \bibinfo{person}{Zhenfei Zhang}, \bibinfo{person}{Yufei Ding}, {and} \bibinfo{person}{Shumo Chu}.} \bibinfo{year}{2021}\natexlab{}.
\newblock \showarticletitle{{ZEN:} Efficient Zero-Knowledge Proofs for Neural Networks}.
\newblock \bibinfo{journal}{\emph{{IACR} Cryptol. ePrint Arch.}} (\bibinfo{year}{2021}), \bibinfo{pages}{87}.
\newblock
\urldef\tempurl%
\url{https://eprint.iacr.org/2021/087}
\showURL{%
\tempurl}


\bibitem[Filecoin(2023)]%
        {ec-gpu}
\bibfield{author}{\bibinfo{person}{Filecoin}.} \bibinfo{year}{2023}\natexlab{}.
\newblock \bibinfo{title}{ec-gpu}.
\newblock \bibinfo{howpublished}{\url{https://github.com/filecoin-project/ec-gpu}}.
\newblock
\newblock
\shownote{Accessed: 2024-01-22}.


\bibitem[Frantar and Alistarh(2023)]%
        {qmoe}
\bibfield{author}{\bibinfo{person}{Elias Frantar} {and} \bibinfo{person}{Dan Alistarh}.} \bibinfo{year}{2023}\natexlab{}.
\newblock \showarticletitle{QMoE: Practical Sub-1-Bit Compression of Trillion-Parameter Models}.
\newblock \bibinfo{journal}{\emph{CoRR}}  \bibinfo{volume}{abs/2310.16795} (\bibinfo{year}{2023}).
\newblock
\urldef\tempurl%
\url{https://doi.org/10.48550/ARXIV.2310.16795}
\showDOI{\tempurl}
\showeprint[arXiv]{2310.16795}


\bibitem[Frantar et~al\mbox{.}(2022)]%
        {gptq}
\bibfield{author}{\bibinfo{person}{Elias Frantar}, \bibinfo{person}{Saleh Ashkboos}, \bibinfo{person}{Torsten Hoefler}, {and} \bibinfo{person}{Dan Alistarh}.} \bibinfo{year}{2022}\natexlab{}.
\newblock \showarticletitle{{GPTQ:} Accurate Post-Training Quantization for Generative Pre-trained Transformers}.
\newblock \bibinfo{journal}{\emph{CoRR}}  \bibinfo{volume}{abs/2210.17323} (\bibinfo{year}{2022}).
\newblock
\urldef\tempurl%
\url{https://doi.org/10.48550/ARXIV.2210.17323}
\showDOI{\tempurl}
\showeprint[arXiv]{2210.17323}


\bibitem[Gabizon and Khovratovich(2022)]%
        {flookup}
\bibfield{author}{\bibinfo{person}{Ariel Gabizon} {and} \bibinfo{person}{Dmitry Khovratovich}.} \bibinfo{year}{2022}\natexlab{}.
\newblock \showarticletitle{flookup: Fractional decomposition-based lookups in quasi-linear time independent of table size}.
\newblock \bibinfo{journal}{\emph{{IACR} Cryptol. ePrint Arch.}} (\bibinfo{year}{2022}), \bibinfo{pages}{1447}.
\newblock
\urldef\tempurl%
\url{https://eprint.iacr.org/2022/1447}
\showURL{%
\tempurl}


\bibitem[Gabizon and Williamson(2020)]%
        {plookup}
\bibfield{author}{\bibinfo{person}{Ariel Gabizon} {and} \bibinfo{person}{Zachary~J. Williamson}.} \bibinfo{year}{2020}\natexlab{}.
\newblock \showarticletitle{plookup: {A} simplified polynomial protocol for lookup tables}.
\newblock \bibinfo{journal}{\emph{{IACR} Cryptol. ePrint Arch.}} (\bibinfo{year}{2020}), \bibinfo{pages}{315}.
\newblock
\urldef\tempurl%
\url{https://eprint.iacr.org/2020/315}
\showURL{%
\tempurl}


\bibitem[Gabizon et~al\mbox{.}(2019)]%
        {plonk}
\bibfield{author}{\bibinfo{person}{Ariel Gabizon}, \bibinfo{person}{Zachary~J. Williamson}, {and} \bibinfo{person}{Oana Ciobotaru}.} \bibinfo{year}{2019}\natexlab{}.
\newblock \showarticletitle{{PLONK:} Permutations over Lagrange-bases for Oecumenical Noninteractive arguments of Knowledge}.
\newblock \bibinfo{journal}{\emph{{IACR} Cryptol. ePrint Arch.}} (\bibinfo{year}{2019}), \bibinfo{pages}{953}.
\newblock
\urldef\tempurl%
\url{https://eprint.iacr.org/2019/953}
\showURL{%
\tempurl}


\bibitem[Garg et~al\mbox{.}(2023)]%
        {zkpot}
\bibfield{author}{\bibinfo{person}{Sanjam Garg}, \bibinfo{person}{Aarushi Goel}, \bibinfo{person}{Somesh Jha}, \bibinfo{person}{Saeed Mahloujifar}, \bibinfo{person}{Mohammad Mahmoody}, \bibinfo{person}{Guru{-}Vamsi Policharla}, {and} \bibinfo{person}{Mingyuan Wang}.} \bibinfo{year}{2023}\natexlab{}.
\newblock \showarticletitle{Experimenting with Zero-Knowledge Proofs of Training}.
\newblock \bibinfo{journal}{\emph{{IACR} Cryptol. ePrint Arch.}} (\bibinfo{year}{2023}), \bibinfo{pages}{1345}.
\newblock
\urldef\tempurl%
\url{https://eprint.iacr.org/2023/1345}
\showURL{%
\tempurl}


\bibitem[Ghodsi et~al\mbox{.}(2017)]%
        {safetynets}
\bibfield{author}{\bibinfo{person}{Zahra Ghodsi}, \bibinfo{person}{Tianyu Gu}, {and} \bibinfo{person}{Siddharth Garg}.} \bibinfo{year}{2017}\natexlab{}.
\newblock \showarticletitle{SafetyNets: Verifiable Execution of Deep Neural Networks on an Untrusted Cloud}. In \bibinfo{booktitle}{\emph{Advances in Neural Information Processing Systems 30: Annual Conference on Neural Information Processing Systems 2017, December 4-9, 2017, Long Beach, CA, {USA}}}, \bibfield{editor}{\bibinfo{person}{Isabelle Guyon}, \bibinfo{person}{Ulrike von Luxburg}, \bibinfo{person}{Samy Bengio}, \bibinfo{person}{Hanna~M. Wallach}, \bibinfo{person}{Rob Fergus}, \bibinfo{person}{S.~V.~N. Vishwanathan}, {and} \bibinfo{person}{Roman Garnett}} (Eds.). \bibinfo{pages}{4672--4681}.
\newblock
\urldef\tempurl%
\url{https://proceedings.neurips.cc/paper/2017/hash/6048ff4e8cb07aa60b6777b6f7384d52-Abstract.html}
\showURL{%
\tempurl}


\bibitem[Goldwasser et~al\mbox{.}(2008)]%
        {gkr}
\bibfield{author}{\bibinfo{person}{Shafi Goldwasser}, \bibinfo{person}{Yael~Tauman Kalai}, {and} \bibinfo{person}{Guy~N. Rothblum}.} \bibinfo{year}{2008}\natexlab{}.
\newblock \showarticletitle{Delegating computation: interactive proofs for muggles}. In \bibinfo{booktitle}{\emph{Proceedings of the 40th Annual {ACM} Symposium on Theory of Computing, Victoria, British Columbia, Canada, May 17-20, 2008}}, \bibfield{editor}{\bibinfo{person}{Cynthia Dwork}} (Ed.). \bibinfo{publisher}{{ACM}}, \bibinfo{pages}{113--122}.
\newblock
\urldef\tempurl%
\url{https://doi.org/10.1145/1374376.1374396}
\showDOI{\tempurl}


\bibitem[Groth(2016)]%
        {groth16}
\bibfield{author}{\bibinfo{person}{Jens Groth}.} \bibinfo{year}{2016}\natexlab{}.
\newblock \showarticletitle{On the Size of Pairing-Based Non-interactive Arguments}. In \bibinfo{booktitle}{\emph{Advances in Cryptology - {EUROCRYPT} 2016 - 35th Annual International Conference on the Theory and Applications of Cryptographic Techniques, Vienna, Austria, May 8-12, 2016, Proceedings, Part {II}}} \emph{(\bibinfo{series}{Lecture Notes in Computer Science}, Vol.~\bibinfo{volume}{9666})}, \bibfield{editor}{\bibinfo{person}{Marc Fischlin} {and} \bibinfo{person}{Jean{-}S{\'{e}}bastien Coron}} (Eds.). \bibinfo{publisher}{Springer}, \bibinfo{pages}{305--326}.
\newblock
\urldef\tempurl%
\url{https://doi.org/10.1007/978-3-662-49896-5\_11}
\showDOI{\tempurl}


\bibitem[Gu et~al\mbox{.}(2022)]%
        {DBLP:journals/corr/abs-2210-07543}
\bibfield{author}{\bibinfo{person}{Chenxi Gu}, \bibinfo{person}{Chengsong Huang}, \bibinfo{person}{Xiaoqing Zheng}, \bibinfo{person}{Kai{-}Wei Chang}, {and} \bibinfo{person}{Cho{-}Jui Hsieh}.} \bibinfo{year}{2022}\natexlab{}.
\newblock \showarticletitle{Watermarking Pre-trained Language Models with Backdooring}.
\newblock \bibinfo{journal}{\emph{CoRR}}  \bibinfo{volume}{abs/2210.07543} (\bibinfo{year}{2022}).
\newblock
\urldef\tempurl%
\url{https://doi.org/10.48550/ARXIV.2210.07543}
\showDOI{\tempurl}
\showeprint[arXiv]{2210.07543}


\bibitem[Hab{\"{o}}ck(2022)]%
        {DBLP:journals/iacr/Habock22a}
\bibfield{author}{\bibinfo{person}{Ulrich Hab{\"{o}}ck}.} \bibinfo{year}{2022}\natexlab{}.
\newblock \showarticletitle{Multivariate lookups based on logarithmic derivatives}.
\newblock \bibinfo{journal}{\emph{{IACR} Cryptol. ePrint Arch.}} (\bibinfo{year}{2022}), \bibinfo{pages}{1530}.
\newblock
\urldef\tempurl%
\url{https://eprint.iacr.org/2022/1530}
\showURL{%
\tempurl}


\bibitem[Hendrycks and Gimpel(2016)]%
        {gelu}
\bibfield{author}{\bibinfo{person}{Dan Hendrycks} {and} \bibinfo{person}{Kevin Gimpel}.} \bibinfo{year}{2016}\natexlab{}.
\newblock \showarticletitle{Bridging Nonlinearities and Stochastic Regularizers with Gaussian Error Linear Units}.
\newblock \bibinfo{journal}{\emph{CoRR}}  \bibinfo{volume}{abs/1606.08415} (\bibinfo{year}{2016}).
\newblock
\showeprint[arXiv]{1606.08415}
\urldef\tempurl%
\url{http://arxiv.org/abs/1606.08415}
\showURL{%
\tempurl}


\bibitem[Hu et~al\mbox{.}(2023)]%
        {DBLP:journals/corr/abs-2310-10669}
\bibfield{author}{\bibinfo{person}{Zhengmian Hu}, \bibinfo{person}{Lichang Chen}, \bibinfo{person}{Xidong Wu}, \bibinfo{person}{Yihan Wu}, \bibinfo{person}{Hongyang Zhang}, {and} \bibinfo{person}{Heng Huang}.} \bibinfo{year}{2023}\natexlab{}.
\newblock \showarticletitle{Unbiased Watermark for Large Language Models}.
\newblock \bibinfo{journal}{\emph{CoRR}}  \bibinfo{volume}{abs/2310.10669} (\bibinfo{year}{2023}).
\newblock
\urldef\tempurl%
\url{https://doi.org/10.48550/ARXIV.2310.10669}
\showDOI{\tempurl}
\showeprint[arXiv]{2310.10669}


\bibitem[Kang et~al\mbox{.}(2022)]%
        {zkml}
\bibfield{author}{\bibinfo{person}{Daniel Kang}, \bibinfo{person}{Tatsunori Hashimoto}, \bibinfo{person}{Ion Stoica}, {and} \bibinfo{person}{Yi Sun}.} \bibinfo{year}{2022}\natexlab{}.
\newblock \showarticletitle{Scaling up Trustless {DNN} Inference with Zero-Knowledge Proofs}.
\newblock \bibinfo{journal}{\emph{CoRR}}  \bibinfo{volume}{abs/2210.08674} (\bibinfo{year}{2022}).
\newblock
\urldef\tempurl%
\url{https://doi.org/10.48550/arXiv.2210.08674}
\showDOI{\tempurl}
\showeprint[arXiv]{2210.08674}


\bibitem[Kirchenbauer et~al\mbox{.}(2023)]%
        {DBLP:conf/icml/KirchenbauerGWK23}
\bibfield{author}{\bibinfo{person}{John Kirchenbauer}, \bibinfo{person}{Jonas Geiping}, \bibinfo{person}{Yuxin Wen}, \bibinfo{person}{Jonathan Katz}, \bibinfo{person}{Ian Miers}, {and} \bibinfo{person}{Tom Goldstein}.} \bibinfo{year}{2023}\natexlab{}.
\newblock \showarticletitle{A Watermark for Large Language Models}. In \bibinfo{booktitle}{\emph{International Conference on Machine Learning, {ICML} 2023, 23-29 July 2023, Honolulu, Hawaii, {USA}}} \emph{(\bibinfo{series}{Proceedings of Machine Learning Research}, Vol.~\bibinfo{volume}{202})}, \bibfield{editor}{\bibinfo{person}{Andreas Krause}, \bibinfo{person}{Emma Brunskill}, \bibinfo{person}{Kyunghyun Cho}, \bibinfo{person}{Barbara Engelhardt}, \bibinfo{person}{Sivan Sabato}, {and} \bibinfo{person}{Jonathan Scarlett}} (Eds.). \bibinfo{publisher}{{PMLR}}, \bibinfo{pages}{17061--17084}.
\newblock
\urldef\tempurl%
\url{https://proceedings.mlr.press/v202/kirchenbauer23a.html}
\showURL{%
\tempurl}


\bibitem[Lee et~al\mbox{.}(2020)]%
        {vcnn}
\bibfield{author}{\bibinfo{person}{Seunghwa Lee}, \bibinfo{person}{Hankyung Ko}, \bibinfo{person}{Jihye Kim}, {and} \bibinfo{person}{Hyunok Oh}.} \bibinfo{year}{2020}\natexlab{}.
\newblock \showarticletitle{{vCNN}: Verifiable Convolutional Neural Network}.
\newblock \bibinfo{journal}{\emph{{IACR} Cryptol. ePrint Arch.}} (\bibinfo{year}{2020}), \bibinfo{pages}{584}.
\newblock
\urldef\tempurl%
\url{https://eprint.iacr.org/2020/584}
\showURL{%
\tempurl}


\bibitem[Ligero et~al\mbox{.}(2019)]%
        {ligero}
\bibfield{author}{\bibinfo{person}{Marta Ligero}, \bibinfo{person}{Guillermo Torres}, \bibinfo{person}{Carles S{\'{a}}nchez}, \bibinfo{person}{Katerine D{\'{\i}}az{-}Chito}, \bibinfo{person}{Raquel Perez}, {and} \bibinfo{person}{Debora Gil}.} \bibinfo{year}{2019}\natexlab{}.
\newblock \showarticletitle{Selection of Radiomics Features based on their Reproducibility}. In \bibinfo{booktitle}{\emph{41st Annual International Conference of the {IEEE} Engineering in Medicine and Biology Society, {EMBC} 2019, Berlin, Germany, July 23-27, 2019}}. \bibinfo{publisher}{{IEEE}}, \bibinfo{pages}{403--408}.
\newblock
\urldef\tempurl%
\url{https://doi.org/10.1109/EMBC.2019.8857879}
\showDOI{\tempurl}


\bibitem[Lin et~al\mbox{.}(2023)]%
        {awq}
\bibfield{author}{\bibinfo{person}{Ji Lin}, \bibinfo{person}{Jiaming Tang}, \bibinfo{person}{Haotian Tang}, \bibinfo{person}{Shang Yang}, \bibinfo{person}{Xingyu Dang}, {and} \bibinfo{person}{Song Han}.} \bibinfo{year}{2023}\natexlab{}.
\newblock \showarticletitle{{AWQ:} Activation-aware Weight Quantization for {LLM} Compression and Acceleration}.
\newblock \bibinfo{journal}{\emph{CoRR}}  \bibinfo{volume}{abs/2306.00978} (\bibinfo{year}{2023}).
\newblock
\urldef\tempurl%
\url{https://doi.org/10.48550/ARXIV.2306.00978}
\showDOI{\tempurl}
\showeprint[arXiv]{2306.00978}


\bibitem[Liu et~al\mbox{.}(2021)]%
        {zkcnn}
\bibfield{author}{\bibinfo{person}{Tianyi Liu}, \bibinfo{person}{Xiang Xie}, {and} \bibinfo{person}{Yupeng Zhang}.} \bibinfo{year}{2021}\natexlab{}.
\newblock \showarticletitle{{zkCNN}: Zero Knowledge Proofs for Convolutional Neural Network Predictions and Accuracy}. In \bibinfo{booktitle}{\emph{{CCS} '21: 2021 {ACM} {SIGSAC} Conference on Computer and Communications Security, Virtual Event, Republic of Korea, November 15 - 19, 2021}}, \bibfield{editor}{\bibinfo{person}{Yongdae Kim}, \bibinfo{person}{Jong Kim}, \bibinfo{person}{Giovanni Vigna}, {and} \bibinfo{person}{Elaine Shi}} (Eds.). \bibinfo{publisher}{{ACM}}, \bibinfo{pages}{2968--2985}.
\newblock
\urldef\tempurl%
\url{https://doi.org/10.1145/3460120.3485379}
\showDOI{\tempurl}


\bibitem[Lund et~al\mbox{.}(1992)]%
        {me}
\bibfield{author}{\bibinfo{person}{Carsten Lund}, \bibinfo{person}{Lance Fortnow}, \bibinfo{person}{Howard~J. Karloff}, {and} \bibinfo{person}{Noam Nisan}.} \bibinfo{year}{1992}\natexlab{}.
\newblock \showarticletitle{Algebraic Methods for Interactive Proof Systems}.
\newblock \bibinfo{journal}{\emph{J. {ACM}}} \bibinfo{volume}{39}, \bibinfo{number}{4} (\bibinfo{year}{1992}), \bibinfo{pages}{859--868}.
\newblock
\urldef\tempurl%
\url{https://doi.org/10.1145/146585.146605}
\showDOI{\tempurl}


\bibitem[Mitsunari(2023)]%
        {mcl}
\bibfield{author}{\bibinfo{person}{Shigeo Mitsunari}.} \bibinfo{year}{2023}\natexlab{}.
\newblock \bibinfo{title}{{MCL}: A portable and fast pairing-based cryptography library}.
\newblock \bibinfo{howpublished}{\url{https://github.com/herumi/mcl}}.
\newblock
\newblock
\shownote{Accessed: 2024-01-22}.


\bibitem[OpenAI(2023)]%
        {gpt4}
\bibfield{author}{\bibinfo{person}{OpenAI}.} \bibinfo{year}{2023}\natexlab{}.
\newblock \showarticletitle{{GPT-4} Technical Report}.
\newblock \bibinfo{journal}{\emph{CoRR}}  \bibinfo{volume}{abs/2303.08774} (\bibinfo{year}{2023}).
\newblock
\urldef\tempurl%
\url{https://doi.org/10.48550/ARXIV.2303.08774}
\showDOI{\tempurl}
\showeprint[arXiv]{2303.08774}


\bibitem[Parno et~al\mbox{.}(2013)]%
        {pinocchio}
\bibfield{author}{\bibinfo{person}{Bryan Parno}, \bibinfo{person}{Jon Howell}, \bibinfo{person}{Craig Gentry}, {and} \bibinfo{person}{Mariana Raykova}.} \bibinfo{year}{2013}\natexlab{}.
\newblock \showarticletitle{Pinocchio: Nearly Practical Verifiable Computation}. In \bibinfo{booktitle}{\emph{2013 {IEEE} Symposium on Security and Privacy, {SP} 2013, Berkeley, CA, USA, May 19-22, 2013}}. \bibinfo{publisher}{{IEEE} Computer Society}, \bibinfo{pages}{238--252}.
\newblock
\urldef\tempurl%
\url{https://doi.org/10.1109/SP.2013.47}
\showDOI{\tempurl}


\bibitem[Pedersen(1991)]%
        {pedersen}
\bibfield{author}{\bibinfo{person}{Torben~P. Pedersen}.} \bibinfo{year}{1991}\natexlab{}.
\newblock \showarticletitle{Non-Interactive and Information-Theoretic Secure Verifiable Secret Sharing}. In \bibinfo{booktitle}{\emph{Advances in Cryptology - {CRYPTO} '91, 11th Annual International Cryptology Conference, Santa Barbara, California, USA, August 11-15, 1991, Proceedings}} \emph{(\bibinfo{series}{Lecture Notes in Computer Science}, Vol.~\bibinfo{volume}{576})}, \bibfield{editor}{\bibinfo{person}{Joan Feigenbaum}} (Ed.). \bibinfo{publisher}{Springer}, \bibinfo{pages}{129--140}.
\newblock
\urldef\tempurl%
\url{https://doi.org/10.1007/3-540-46766-1\_9}
\showDOI{\tempurl}


\bibitem[Posen and Kattis(2022)]%
        {caulk+}
\bibfield{author}{\bibinfo{person}{Jim Posen} {and} \bibinfo{person}{Assimakis~A. Kattis}.} \bibinfo{year}{2022}\natexlab{}.
\newblock \showarticletitle{Caulk+: Table-independent lookup arguments}.
\newblock \bibinfo{journal}{\emph{{IACR} Cryptol. ePrint Arch.}} (\bibinfo{year}{2022}), \bibinfo{pages}{957}.
\newblock
\urldef\tempurl%
\url{https://eprint.iacr.org/2022/957}
\showURL{%
\tempurl}


\bibitem[Raffel et~al\mbox{.}(2020)]%
        {c4}
\bibfield{author}{\bibinfo{person}{Colin Raffel}, \bibinfo{person}{Noam Shazeer}, \bibinfo{person}{Adam Roberts}, \bibinfo{person}{Katherine Lee}, \bibinfo{person}{Sharan Narang}, \bibinfo{person}{Michael Matena}, \bibinfo{person}{Yanqi Zhou}, \bibinfo{person}{Wei Li}, {and} \bibinfo{person}{Peter~J. Liu}.} \bibinfo{year}{2020}\natexlab{}.
\newblock \showarticletitle{Exploring the Limits of Transfer Learning with a Unified Text-to-Text Transformer}.
\newblock \bibinfo{journal}{\emph{J. Mach. Learn. Res.}}  \bibinfo{volume}{21} (\bibinfo{year}{2020}), \bibinfo{pages}{140:1--140:67}.
\newblock
\urldef\tempurl%
\url{http://jmlr.org/papers/v21/20-074.html}
\showURL{%
\tempurl}


\bibitem[Schwartz(1980)]%
        {DBLP:journals/jacm/Schwartz80}
\bibfield{author}{\bibinfo{person}{Jacob~T. Schwartz}.} \bibinfo{year}{1980}\natexlab{}.
\newblock \showarticletitle{Fast Probabilistic Algorithms for Verification of Polynomial Identities}.
\newblock \bibinfo{journal}{\emph{J. {ACM}}} \bibinfo{volume}{27}, \bibinfo{number}{4} (\bibinfo{year}{1980}), \bibinfo{pages}{701--717}.
\newblock
\urldef\tempurl%
\url{https://doi.org/10.1145/322217.322225}
\showDOI{\tempurl}


\bibitem[Shazeer(2020)]%
        {swiglu}
\bibfield{author}{\bibinfo{person}{Noam Shazeer}.} \bibinfo{year}{2020}\natexlab{}.
\newblock \showarticletitle{{GLU} Variants Improve Transformer}.
\newblock \bibinfo{journal}{\emph{CoRR}}  \bibinfo{volume}{abs/2002.05202} (\bibinfo{year}{2020}).
\newblock
\showeprint[arXiv]{2002.05202}
\urldef\tempurl%
\url{https://arxiv.org/abs/2002.05202}
\showURL{%
\tempurl}


\bibitem[Su et~al\mbox{.}(2024)]%
        {rope}
\bibfield{author}{\bibinfo{person}{Jianlin Su}, \bibinfo{person}{Murtadha H.~M. Ahmed}, \bibinfo{person}{Yu Lu}, \bibinfo{person}{Shengfeng Pan}, \bibinfo{person}{Wen Bo}, {and} \bibinfo{person}{Yunfeng Liu}.} \bibinfo{year}{2024}\natexlab{}.
\newblock \showarticletitle{RoFormer: Enhanced transformer with Rotary Position Embedding}.
\newblock \bibinfo{journal}{\emph{Neurocomputing}}  \bibinfo{volume}{568} (\bibinfo{year}{2024}), \bibinfo{pages}{127063}.
\newblock
\urldef\tempurl%
\url{https://doi.org/10.1016/J.NEUCOM.2023.127063}
\showDOI{\tempurl}


\bibitem[Thaler(2013)]%
        {DBLP:conf/crypto/Thaler13}
\bibfield{author}{\bibinfo{person}{Justin Thaler}.} \bibinfo{year}{2013}\natexlab{}.
\newblock \showarticletitle{Time-Optimal Interactive Proofs for Circuit Evaluation}. In \bibinfo{booktitle}{\emph{Advances in Cryptology - {CRYPTO} 2013 - 33rd Annual Cryptology Conference, Santa Barbara, CA, USA, August 18-22, 2013. Proceedings, Part {II}}} \emph{(\bibinfo{series}{Lecture Notes in Computer Science}, Vol.~\bibinfo{volume}{8043})}, \bibfield{editor}{\bibinfo{person}{Ran Canetti} {and} \bibinfo{person}{Juan~A. Garay}} (Eds.). \bibinfo{publisher}{Springer}, \bibinfo{pages}{71--89}.
\newblock
\urldef\tempurl%
\url{https://doi.org/10.1007/978-3-642-40084-1\_5}
\showDOI{\tempurl}


\bibitem[Thaler(2022)]%
        {DBLP:journals/ftsec/Thaler22}
\bibfield{author}{\bibinfo{person}{Justin Thaler}.} \bibinfo{year}{2022}\natexlab{}.
\newblock \showarticletitle{Proofs, Arguments, and Zero-Knowledge}.
\newblock \bibinfo{journal}{\emph{Found. Trends Priv. Secur.}} \bibinfo{volume}{4}, \bibinfo{number}{2-4} (\bibinfo{year}{2022}), \bibinfo{pages}{117--660}.
\newblock
\urldef\tempurl%
\url{https://doi.org/10.1561/3300000030}
\showDOI{\tempurl}


\bibitem[Touvron et~al\mbox{.}(2023a)]%
        {llama}
\bibfield{author}{\bibinfo{person}{Hugo Touvron}, \bibinfo{person}{Thibaut Lavril}, \bibinfo{person}{Gautier Izacard}, \bibinfo{person}{Xavier Martinet}, \bibinfo{person}{Marie{-}Anne Lachaux}, \bibinfo{person}{Timoth{\'{e}}e Lacroix}, \bibinfo{person}{Baptiste Rozi{\`{e}}re}, \bibinfo{person}{Naman Goyal}, \bibinfo{person}{Eric Hambro}, \bibinfo{person}{Faisal Azhar}, \bibinfo{person}{Aur{\'{e}}lien Rodriguez}, \bibinfo{person}{Armand Joulin}, \bibinfo{person}{Edouard Grave}, {and} \bibinfo{person}{Guillaume Lample}.} \bibinfo{year}{2023}\natexlab{a}.
\newblock \showarticletitle{LLaMA: Open and Efficient Foundation Language Models}.
\newblock \bibinfo{journal}{\emph{CoRR}}  \bibinfo{volume}{abs/2302.13971} (\bibinfo{year}{2023}).
\newblock
\urldef\tempurl%
\url{https://doi.org/10.48550/ARXIV.2302.13971}
\showDOI{\tempurl}
\showeprint[arXiv]{2302.13971}


\bibitem[Touvron et~al\mbox{.}(2023b)]%
        {llama2}
\bibfield{author}{\bibinfo{person}{Hugo Touvron}, \bibinfo{person}{Louis Martin}, \bibinfo{person}{Kevin Stone}, \bibinfo{person}{Peter Albert}, \bibinfo{person}{Amjad Almahairi}, \bibinfo{person}{Yasmine Babaei}, \bibinfo{person}{Nikolay Bashlykov}, \bibinfo{person}{Soumya Batra}, \bibinfo{person}{Prajjwal Bhargava}, \bibinfo{person}{Shruti Bhosale}, \bibinfo{person}{Dan Bikel}, \bibinfo{person}{Lukas Blecher}, \bibinfo{person}{Cristian Canton{-}Ferrer}, \bibinfo{person}{Moya Chen}, \bibinfo{person}{Guillem Cucurull}, \bibinfo{person}{David Esiobu}, \bibinfo{person}{Jude Fernandes}, \bibinfo{person}{Jeremy Fu}, \bibinfo{person}{Wenyin Fu}, \bibinfo{person}{Brian Fuller}, \bibinfo{person}{Cynthia Gao}, \bibinfo{person}{Vedanuj Goswami}, \bibinfo{person}{Naman Goyal}, \bibinfo{person}{Anthony Hartshorn}, \bibinfo{person}{Saghar Hosseini}, \bibinfo{person}{Rui Hou}, \bibinfo{person}{Hakan Inan}, \bibinfo{person}{Marcin Kardas}, \bibinfo{person}{Viktor Kerkez}, \bibinfo{person}{Madian Khabsa},
  \bibinfo{person}{Isabel Kloumann}, \bibinfo{person}{Artem Korenev}, \bibinfo{person}{Punit~Singh Koura}, \bibinfo{person}{Marie{-}Anne Lachaux}, \bibinfo{person}{Thibaut Lavril}, \bibinfo{person}{Jenya Lee}, \bibinfo{person}{Diana Liskovich}, \bibinfo{person}{Yinghai Lu}, \bibinfo{person}{Yuning Mao}, \bibinfo{person}{Xavier Martinet}, \bibinfo{person}{Todor Mihaylov}, \bibinfo{person}{Pushkar Mishra}, \bibinfo{person}{Igor Molybog}, \bibinfo{person}{Yixin Nie}, \bibinfo{person}{Andrew Poulton}, \bibinfo{person}{Jeremy Reizenstein}, \bibinfo{person}{Rashi Rungta}, \bibinfo{person}{Kalyan Saladi}, \bibinfo{person}{Alan Schelten}, \bibinfo{person}{Ruan Silva}, \bibinfo{person}{Eric~Michael Smith}, \bibinfo{person}{Ranjan Subramanian}, \bibinfo{person}{Xiaoqing~Ellen Tan}, \bibinfo{person}{Binh Tang}, \bibinfo{person}{Ross Taylor}, \bibinfo{person}{Adina Williams}, \bibinfo{person}{Jian~Xiang Kuan}, \bibinfo{person}{Puxin Xu}, \bibinfo{person}{Zheng Yan}, \bibinfo{person}{Iliyan Zarov}, \bibinfo{person}{Yuchen
  Zhang}, \bibinfo{person}{Angela Fan}, \bibinfo{person}{Melanie Kambadur}, \bibinfo{person}{Sharan Narang}, \bibinfo{person}{Aur{\'{e}}lien Rodriguez}, \bibinfo{person}{Robert Stojnic}, \bibinfo{person}{Sergey Edunov}, {and} \bibinfo{person}{Thomas Scialom}.} \bibinfo{year}{2023}\natexlab{b}.
\newblock \showarticletitle{Llama 2: Open Foundation and Fine-Tuned Chat Models}.
\newblock \bibinfo{journal}{\emph{CoRR}}  \bibinfo{volume}{abs/2307.09288} (\bibinfo{year}{2023}).
\newblock
\urldef\tempurl%
\url{https://doi.org/10.48550/ARXIV.2307.09288}
\showDOI{\tempurl}
\showeprint[arXiv]{2307.09288}


\bibitem[Vaswani et~al\mbox{.}(2017)]%
        {attention}
\bibfield{author}{\bibinfo{person}{Ashish Vaswani}, \bibinfo{person}{Noam Shazeer}, \bibinfo{person}{Niki Parmar}, \bibinfo{person}{Jakob Uszkoreit}, \bibinfo{person}{Llion Jones}, \bibinfo{person}{Aidan~N. Gomez}, \bibinfo{person}{Lukasz Kaiser}, {and} \bibinfo{person}{Illia Polosukhin}.} \bibinfo{year}{2017}\natexlab{}.
\newblock \showarticletitle{Attention is All you Need}. In \bibinfo{booktitle}{\emph{Advances in Neural Information Processing Systems 30: Annual Conference on Neural Information Processing Systems 2017, December 4-9, 2017, Long Beach, CA, {USA}}}, \bibfield{editor}{\bibinfo{person}{Isabelle Guyon}, \bibinfo{person}{Ulrike von Luxburg}, \bibinfo{person}{Samy Bengio}, \bibinfo{person}{Hanna~M. Wallach}, \bibinfo{person}{Rob Fergus}, \bibinfo{person}{S.~V.~N. Vishwanathan}, {and} \bibinfo{person}{Roman Garnett}} (Eds.). \bibinfo{pages}{5998--6008}.
\newblock
\urldef\tempurl%
\url{https://proceedings.neurips.cc/paper/2017/hash/3f5ee243547dee91fbd053c1c4a845aa-Abstract.html}
\showURL{%
\tempurl}


\bibitem[Wahby et~al\mbox{.}(2018)]%
        {hyrax}
\bibfield{author}{\bibinfo{person}{Riad~S. Wahby}, \bibinfo{person}{Ioanna Tzialla}, \bibinfo{person}{Abhi Shelat}, \bibinfo{person}{Justin Thaler}, {and} \bibinfo{person}{Michael Walfish}.} \bibinfo{year}{2018}\natexlab{}.
\newblock \showarticletitle{Doubly-Efficient {zkSNARKs} Without Trusted Setup}. In \bibinfo{booktitle}{\emph{2018 {IEEE} Symposium on Security and Privacy, {SP} 2018, Proceedings, 21-23 May 2018, San Francisco, California, {USA}}}. \bibinfo{publisher}{{IEEE} Computer Society}, \bibinfo{pages}{926--943}.
\newblock
\urldef\tempurl%
\url{https://doi.org/10.1109/SP.2018.00060}
\showDOI{\tempurl}


\bibitem[Wang and Hoang(2023)]%
        {ezdps}
\bibfield{author}{\bibinfo{person}{Haodi Wang} {and} \bibinfo{person}{Thang Hoang}.} \bibinfo{year}{2023}\natexlab{}.
\newblock \showarticletitle{{ezDPS}: An Efficient and Zero-Knowledge Machine Learning Inference Pipeline}.
\newblock \bibinfo{journal}{\emph{Proc. Priv. Enhancing Technol.}} \bibinfo{volume}{2023}, \bibinfo{number}{2} (\bibinfo{year}{2023}), \bibinfo{pages}{430--448}.
\newblock
\urldef\tempurl%
\url{https://doi.org/10.56553/popets-2023-0061}
\showDOI{\tempurl}


\bibitem[Weng et~al\mbox{.}(2021)]%
        {mystique}
\bibfield{author}{\bibinfo{person}{Chenkai Weng}, \bibinfo{person}{Kang Yang}, \bibinfo{person}{Xiang Xie}, \bibinfo{person}{Jonathan Katz}, {and} \bibinfo{person}{Xiao Wang}.} \bibinfo{year}{2021}\natexlab{}.
\newblock \showarticletitle{{Mystique}: Efficient Conversions for Zero-Knowledge Proofs with Applications to Machine Learning}. In \bibinfo{booktitle}{\emph{30th {USENIX} Security Symposium, {USENIX} Security 2021, August 11-13, 2021}}, \bibfield{editor}{\bibinfo{person}{Michael Bailey} {and} \bibinfo{person}{Rachel Greenstadt}} (Eds.). \bibinfo{publisher}{{USENIX} Association}, \bibinfo{pages}{501--518}.
\newblock
\urldef\tempurl%
\url{https://www.usenix.org/conference/usenixsecurity21/presentation/weng}
\showURL{%
\tempurl}


\bibitem[Weng et~al\mbox{.}(2023)]%
        {pvcnn}
\bibfield{author}{\bibinfo{person}{Jia{-}Si Weng}, \bibinfo{person}{Jian Weng}, \bibinfo{person}{Gui Tang}, \bibinfo{person}{Anjia Yang}, \bibinfo{person}{Ming Li}, {and} \bibinfo{person}{Jia{-}Nan Liu}.} \bibinfo{year}{2023}\natexlab{}.
\newblock \showarticletitle{pvCNN: Privacy-Preserving and Verifiable Convolutional Neural Network Testing}.
\newblock \bibinfo{journal}{\emph{{IEEE} Trans. Inf. Forensics Secur.}}  \bibinfo{volume}{18} (\bibinfo{year}{2023}), \bibinfo{pages}{2218--2233}.
\newblock
\urldef\tempurl%
\url{https://doi.org/10.1109/TIFS.2023.3262932}
\showDOI{\tempurl}


\bibitem[Weng et~al\mbox{.}(2022)]%
        {poul}
\bibfield{author}{\bibinfo{person}{Jiasi Weng}, \bibinfo{person}{Shenglong Yao}, \bibinfo{person}{Yuefeng Du}, \bibinfo{person}{Junjie Huang}, \bibinfo{person}{Jian Weng}, {and} \bibinfo{person}{Cong Wang}.} \bibinfo{year}{2022}\natexlab{}.
\newblock \showarticletitle{Proof of Unlearning: Definitions and Instantiation}.
\newblock \bibinfo{journal}{\emph{CoRR}}  \bibinfo{volume}{abs/2210.11334} (\bibinfo{year}{2022}).
\newblock
\urldef\tempurl%
\url{https://doi.org/10.48550/ARXIV.2210.11334}
\showDOI{\tempurl}
\showeprint[arXiv]{2210.11334}


\bibitem[Xiao et~al\mbox{.}(2023)]%
        {smoothquant}
\bibfield{author}{\bibinfo{person}{Guangxuan Xiao}, \bibinfo{person}{Ji Lin}, \bibinfo{person}{Micka{\"{e}}l Seznec}, \bibinfo{person}{Hao Wu}, \bibinfo{person}{Julien Demouth}, {and} \bibinfo{person}{Song Han}.} \bibinfo{year}{2023}\natexlab{}.
\newblock \showarticletitle{SmoothQuant: Accurate and Efficient Post-Training Quantization for Large Language Models}. In \bibinfo{booktitle}{\emph{International Conference on Machine Learning, {ICML} 2023, 23-29 July 2023, Honolulu, Hawaii, {USA}}} \emph{(\bibinfo{series}{Proceedings of Machine Learning Research}, Vol.~\bibinfo{volume}{202})}, \bibfield{editor}{\bibinfo{person}{Andreas Krause}, \bibinfo{person}{Emma Brunskill}, \bibinfo{person}{Kyunghyun Cho}, \bibinfo{person}{Barbara Engelhardt}, \bibinfo{person}{Sivan Sabato}, {and} \bibinfo{person}{Jonathan Scarlett}} (Eds.). \bibinfo{publisher}{{PMLR}}, \bibinfo{pages}{38087--38099}.
\newblock
\urldef\tempurl%
\url{https://proceedings.mlr.press/v202/xiao23c.html}
\showURL{%
\tempurl}


\bibitem[Xie et~al\mbox{.}(2019)]%
        {libra}
\bibfield{author}{\bibinfo{person}{Tiancheng Xie}, \bibinfo{person}{Jiaheng Zhang}, \bibinfo{person}{Yupeng Zhang}, \bibinfo{person}{Charalampos Papamanthou}, {and} \bibinfo{person}{Dawn Song}.} \bibinfo{year}{2019}\natexlab{}.
\newblock \showarticletitle{Libra: Succinct Zero-Knowledge Proofs with Optimal Prover Computation}. In \bibinfo{booktitle}{\emph{Advances in Cryptology - {CRYPTO} 2019 - 39th Annual International Cryptology Conference, Santa Barbara, CA, USA, August 18-22, 2019, Proceedings, Part {III}}} \emph{(\bibinfo{series}{Lecture Notes in Computer Science}, Vol.~\bibinfo{volume}{11694})}, \bibfield{editor}{\bibinfo{person}{Alexandra Boldyreva} {and} \bibinfo{person}{Daniele Micciancio}} (Eds.). \bibinfo{publisher}{Springer}, \bibinfo{pages}{733--764}.
\newblock
\urldef\tempurl%
\url{https://doi.org/10.1007/978-3-030-26954-8\_24}
\showDOI{\tempurl}


\bibitem[Xie et~al\mbox{.}(2022)]%
        {orion}
\bibfield{author}{\bibinfo{person}{Tiancheng Xie}, \bibinfo{person}{Yupeng Zhang}, {and} \bibinfo{person}{Dawn Song}.} \bibinfo{year}{2022}\natexlab{}.
\newblock \showarticletitle{Orion: Zero Knowledge Proof with Linear Prover Time}. In \bibinfo{booktitle}{\emph{Advances in Cryptology - {CRYPTO} 2022 - 42nd Annual International Cryptology Conference, {CRYPTO} 2022, Santa Barbara, CA, USA, August 15-18, 2022, Proceedings, Part {IV}}} \emph{(\bibinfo{series}{Lecture Notes in Computer Science}, Vol.~\bibinfo{volume}{13510})}, \bibfield{editor}{\bibinfo{person}{Yevgeniy Dodis} {and} \bibinfo{person}{Thomas Shrimpton}} (Eds.). \bibinfo{publisher}{Springer}, \bibinfo{pages}{299--328}.
\newblock
\urldef\tempurl%
\url{https://doi.org/10.1007/978-3-031-15985-5\_11}
\showDOI{\tempurl}


\bibitem[Zapico et~al\mbox{.}(2022a)]%
        {caulk}
\bibfield{author}{\bibinfo{person}{Arantxa Zapico}, \bibinfo{person}{Vitalik Buterin}, \bibinfo{person}{Dmitry Khovratovich}, \bibinfo{person}{Mary Maller}, \bibinfo{person}{Anca Nitulescu}, {and} \bibinfo{person}{Mark Simkin}.} \bibinfo{year}{2022}\natexlab{a}.
\newblock \showarticletitle{Caulk: Lookup Arguments in Sublinear Time}. In \bibinfo{booktitle}{\emph{Proceedings of the 2022 {ACM} {SIGSAC} Conference on Computer and Communications Security, {CCS} 2022, Los Angeles, CA, USA, November 7-11, 2022}}, \bibfield{editor}{\bibinfo{person}{Heng Yin}, \bibinfo{person}{Angelos Stavrou}, \bibinfo{person}{Cas Cremers}, {and} \bibinfo{person}{Elaine Shi}} (Eds.). \bibinfo{publisher}{{ACM}}, \bibinfo{pages}{3121--3134}.
\newblock
\urldef\tempurl%
\url{https://doi.org/10.1145/3548606.3560646}
\showDOI{\tempurl}


\bibitem[Zapico et~al\mbox{.}(2022b)]%
        {baloo}
\bibfield{author}{\bibinfo{person}{Arantxa Zapico}, \bibinfo{person}{Ariel Gabizon}, \bibinfo{person}{Dmitry Khovratovich}, \bibinfo{person}{Mary Maller}, {and} \bibinfo{person}{Carla R{\`{a}}fols}.} \bibinfo{year}{2022}\natexlab{b}.
\newblock \showarticletitle{Baloo: Nearly Optimal Lookup Arguments}.
\newblock \bibinfo{journal}{\emph{{IACR} Cryptol. ePrint Arch.}} (\bibinfo{year}{2022}), \bibinfo{pages}{1565}.
\newblock
\urldef\tempurl%
\url{https://eprint.iacr.org/2022/1565}
\showURL{%
\tempurl}


\bibitem[Zhang and Sennrich(2019)]%
        {rmsnorm}
\bibfield{author}{\bibinfo{person}{Biao Zhang} {and} \bibinfo{person}{Rico Sennrich}.} \bibinfo{year}{2019}\natexlab{}.
\newblock \showarticletitle{Root Mean Square Layer Normalization}. In \bibinfo{booktitle}{\emph{Advances in Neural Information Processing Systems 32: Annual Conference on Neural Information Processing Systems 2019, NeurIPS 2019, December 8-14, 2019, Vancouver, BC, Canada}}, \bibfield{editor}{\bibinfo{person}{Hanna~M. Wallach}, \bibinfo{person}{Hugo Larochelle}, \bibinfo{person}{Alina Beygelzimer}, \bibinfo{person}{Florence d'Alch{\'{e}}{-}Buc}, \bibinfo{person}{Emily~B. Fox}, {and} \bibinfo{person}{Roman Garnett}} (Eds.). \bibinfo{pages}{12360--12371}.
\newblock
\urldef\tempurl%
\url{https://proceedings.neurips.cc/paper/2019/hash/1e8a19426224ca89e83cef47f1e7f53b-Abstract.html}
\showURL{%
\tempurl}


\bibitem[Zhang et~al\mbox{.}(2020)]%
        {DBLP:conf/ccs/ZhangFZS20}
\bibfield{author}{\bibinfo{person}{Jiaheng Zhang}, \bibinfo{person}{Zhiyong Fang}, \bibinfo{person}{Yupeng Zhang}, {and} \bibinfo{person}{Dawn Song}.} \bibinfo{year}{2020}\natexlab{}.
\newblock \showarticletitle{Zero Knowledge Proofs for Decision Tree Predictions and Accuracy}. In \bibinfo{booktitle}{\emph{{CCS} '20: 2020 {ACM} {SIGSAC} Conference on Computer and Communications Security, Virtual Event, USA, November 9-13, 2020}}, \bibfield{editor}{\bibinfo{person}{Jay Ligatti}, \bibinfo{person}{Xinming Ou}, \bibinfo{person}{Jonathan Katz}, {and} \bibinfo{person}{Giovanni Vigna}} (Eds.). \bibinfo{publisher}{{ACM}}, \bibinfo{pages}{2039--2053}.
\newblock
\urldef\tempurl%
\url{https://doi.org/10.1145/3372297.3417278}
\showDOI{\tempurl}


\bibitem[Zhang et~al\mbox{.}(2021)]%
        {DBLP:conf/ccs/ZhangLWZSXZ21}
\bibfield{author}{\bibinfo{person}{Jiaheng Zhang}, \bibinfo{person}{Tianyi Liu}, \bibinfo{person}{Weijie Wang}, \bibinfo{person}{Yinuo Zhang}, \bibinfo{person}{Dawn Song}, \bibinfo{person}{Xiang Xie}, {and} \bibinfo{person}{Yupeng Zhang}.} \bibinfo{year}{2021}\natexlab{}.
\newblock \showarticletitle{Doubly Efficient Interactive Proofs for General Arithmetic Circuits with Linear Prover Time}. In \bibinfo{booktitle}{\emph{{CCS} '21: 2021 {ACM} {SIGSAC} Conference on Computer and Communications Security, Virtual Event, Republic of Korea, November 15 - 19, 2021}}, \bibfield{editor}{\bibinfo{person}{Yongdae Kim}, \bibinfo{person}{Jong Kim}, \bibinfo{person}{Giovanni Vigna}, {and} \bibinfo{person}{Elaine Shi}} (Eds.). \bibinfo{publisher}{{ACM}}, \bibinfo{pages}{159--177}.
\newblock
\urldef\tempurl%
\url{https://doi.org/10.1145/3460120.3484767}
\showDOI{\tempurl}


\bibitem[Zhang et~al\mbox{.}(2022)]%
        {opt}
\bibfield{author}{\bibinfo{person}{Susan Zhang}, \bibinfo{person}{Stephen Roller}, \bibinfo{person}{Naman Goyal}, \bibinfo{person}{Mikel Artetxe}, \bibinfo{person}{Moya Chen}, \bibinfo{person}{Shuohui Chen}, \bibinfo{person}{Christopher Dewan}, \bibinfo{person}{Mona~T. Diab}, \bibinfo{person}{Xian Li}, \bibinfo{person}{Xi~Victoria Lin}, \bibinfo{person}{Todor Mihaylov}, \bibinfo{person}{Myle Ott}, \bibinfo{person}{Sam Shleifer}, \bibinfo{person}{Kurt Shuster}, \bibinfo{person}{Daniel Simig}, \bibinfo{person}{Punit~Singh Koura}, \bibinfo{person}{Anjali Sridhar}, \bibinfo{person}{Tianlu Wang}, {and} \bibinfo{person}{Luke Zettlemoyer}.} \bibinfo{year}{2022}\natexlab{}.
\newblock \showarticletitle{{OPT:} Open Pre-trained Transformer Language Models}.
\newblock \bibinfo{journal}{\emph{CoRR}}  \bibinfo{volume}{abs/2205.01068} (\bibinfo{year}{2022}).
\newblock
\urldef\tempurl%
\url{https://doi.org/10.48550/ARXIV.2205.01068}
\showDOI{\tempurl}
\showeprint[arXiv]{2205.01068}


\bibitem[Zhao et~al\mbox{.}(2021)]%
        {veriml}
\bibfield{author}{\bibinfo{person}{Lingchen Zhao}, \bibinfo{person}{Qian Wang}, \bibinfo{person}{Cong Wang}, \bibinfo{person}{Qi Li}, \bibinfo{person}{Chao Shen}, {and} \bibinfo{person}{Bo Feng}.} \bibinfo{year}{2021}\natexlab{}.
\newblock \showarticletitle{VeriML: Enabling Integrity Assurances and Fair Payments for Machine Learning as a Service}.
\newblock \bibinfo{journal}{\emph{{IEEE} Trans. Parallel Distributed Syst.}} \bibinfo{volume}{32}, \bibinfo{number}{10} (\bibinfo{year}{2021}), \bibinfo{pages}{2524--2540}.
\newblock
\urldef\tempurl%
\url{https://doi.org/10.1109/TPDS.2021.3068195}
\showDOI{\tempurl}


\bibitem[Zippel(1979)]%
        {DBLP:conf/eurosam/Zippel79}
\bibfield{author}{\bibinfo{person}{Richard Zippel}.} \bibinfo{year}{1979}\natexlab{}.
\newblock \showarticletitle{Probabilistic algorithms for sparse polynomials}. In \bibinfo{booktitle}{\emph{Symbolic and Algebraic Computation, {EUROSAM} '79, An International Symposiumon Symbolic and Algebraic Computation, Marseille, France, June 1979, Proceedings}} \emph{(\bibinfo{series}{Lecture Notes in Computer Science}, Vol.~\bibinfo{volume}{72})}, \bibfield{editor}{\bibinfo{person}{Edward~W. Ng}} (Ed.). \bibinfo{publisher}{Springer}, \bibinfo{pages}{216--226}.
\newblock
\urldef\tempurl%
\url{https://doi.org/10.1007/3-540-09519-5\_73}
\showDOI{\tempurl}


\end{thebibliography}

\appendix
\section{Code} Our open-source implementation is available at \url{https://github.com/jvhs0706/zkllm-ccs2024}.

\section{Additional related works} \label{appendix:related-works}

In this appendix, we overview of two classes of related studies, and explain their relatedness and distinction with this study.

\medskip
\noindent\textbf{LLM quantization.} To facilitate the deployment of cryptographic tools, this study employs quantization to map the entire computation over LLMs into finite fields. Despite state-of-the-art quantization methods  (e.g., \tt{LLM.int8()}\cite{llmint8}, SmoothQuant \cite{smoothquant}, GPTQ\cite{gptq}, AWQ\cite{awq}, QuIP \cite{quip}, QMoE\cite{qmoe}, QLoRA\cite{qlora}) already achieving low-bit compression of model parameters and activations in LLMs, intermediate computations between these values still involve floating-point numbers to maintain accuracy under low-bit settings. Consequently, these quantization methods are not directly applicable to our study, where computations are strictly confined to finite fields. In our experiments, we utilize a larger bit size (16 bits) to ensure the preservation of accuracy within the finite field constraints.

\medskip
\noindent\textbf{LLM watermark.} Watermarks in LLMs \cite{DBLP:conf/icml/KirchenbauerGWK23,DBLP:journals/corr/abs-2310-10669,DBLP:journals/corr/abs-2210-07543} serve as an additional mechanism to indicate that an output is generated by a specific model. However, it is crucial to note that watermarks in LLMs have vastly different application domains compared to zkLLM. The primary distinction lies in the nature of the assurances they provide. Watermarks offer a form of model identification, while zkLLM, in contrast, asserts a much stronger statement by proving the correctness of the output concerning the committed parameters and the prescribed computational process within the LLM. Unlike watermarks, zkLLM provides a guarantee that the computational process has not been tampered with, offering a more robust and verifiable assurance of the authenticity of the output.

\section{Appendix of Section \ref{sec:analysis}}
\subsection{Appendix on Error Analysis}\label{appendix:error-anal}

In this appendix, we analyze the numerical of \tt{zkAttn}, by giving a detailed proof of Theorem \ref{thm:error-bound}.
\begin{proof}[Proof of Theorem \ref{thm:error-bound}]
    
We first consider the errors introduced in the exponentiation operation, with the input of $-B < z\leq 0$, such that $-z$ is decomposed as $\left(x^{(0)}, x^{(1)}, \dots, x^{(K-1)}\right)$ following \eqref{eq:decomp}. The error incurred in the estimation of $\exp\left(\frac{z}{\gamma\sqrt{d}}\right)$ can be expressed as \begin{align}\label{eq:error-exp}
    \abs{\exp\left(\frac{z}{\gamma\sqrt{d}}\right) - \frac{1}{\theta} \prod_{k=0}^{K-1}y^{(k)}},
\end{align}where each $(x^{(k)}, y^{(k)})\in \mathbf{T}^{(k)}$ is encoded in the lookup tables $\mathbf{T}^{(k)}$s after the optimization for the most and least significant segments in Section \ref{sec:zkattn-optimization}.

For $z\leq -B_{K-M}$, \eqref{eq:error-exp} becomes $\exp\left(\frac{z}{\gamma d}\right)$ as at least one of the $y^{(k)}$s become $0$. Therefore, we focus on the case that $-B_{K-M} < z \leq 0$, such that $-z$ is decompsed as $(x_0, x_1, \dots, x_{K-M -1}, 0, 0, \dots, 0)$, \eqref{eq:error-exp} becomes 
\begin{small}
    \begin{align}
    ~ & \frac{1}{\theta}\abs{\prod_{k=0}^{K-M-1}\theta^{(k)}\exp\left(-\frac{B^{(k)}}{\gamma \sqrt{d}}x^{(k)}\right) - \prod_{k=L}^{K-M-1}\round{\theta^{(k)}\exp\left(-\frac{B^{(k)}}{\gamma \sqrt{d}}x^{(k)}\right)}} \\
    \leq & \frac{1}{\theta}\left(\abs{\prod_{k=0}^{K-M-1}\theta^{(k)}\exp\left(-\frac{B^{(k)}}{\gamma \sqrt{d}}x^{(k)}\right) - \prod_{k=L}^{K-M-1}\theta^{(k)}\exp\left(-\frac{B^{(k)}}{\gamma \sqrt{d}}x^{(k)}\right)}\right. + \nonumber\\ 
    ~ & \left.\abs{\prod_{k=L}^{K-M-1}\theta^{(k)}\exp\left(-\frac{B^{(k)}}{\gamma \sqrt{d}}x^{(k)}\right) - \prod_{k=L}^{K-M-1}\round{\theta^{(k)}\exp\left(-\frac{B^{(k)}}{\gamma \sqrt{d}}x^{(k)}\right)}}\right) \\
    \leq &\abs{\prod_{k=0}^{K-M-1}\exp\left(-\frac{B^{(k)}}{\gamma \sqrt{d}}x^{(k)}\right) - \prod_{k=L}^{K-M-1}\exp\left(-\frac{B^{(k)}}{\gamma \sqrt{d}}x^{(k)}\right)}  \nonumber\\ 
    ~ & + \frac{1}{\theta}\abs{\prod_{k=L}^{K-M-1}\theta^{(k)}\exp\left(-\frac{B^{(k)}}{\gamma \sqrt{d}}x^{(k)}\right) - \prod_{k=L}^{K-M-1}\round{\theta^{(k)}\exp\left(-\frac{B^{(k)}}{\gamma \sqrt{d}}x^{(k)}\right)}}, \label{eq:error-exp-decomposed}
\end{align}
\end{small}
 where the first term of \eqref{eq:error-exp-decomposed} can be upper bounded by \begin{equation}
    \exp\left(\frac{z}{\gamma\sqrt{d}}\right) \left(\exp\left(\frac{B_L}{\gamma\sqrt{d}}\right) - 1\right). 
\end{equation}Therefore, we focus our effort on bounding the second term, where \begin{small}
    \begin{align}
    ~ & \frac{1}{\theta}\abs{\prod_{k=L}^{K-M-1}\theta^{(k)}\exp\left(-\frac{B^{(k)}}{\gamma \sqrt{d}}x^{(k)}\right) - \prod_{k=L}^{K-M-1}\round{\theta^{(k)}\exp\left(-\frac{B^{(k)}}{\gamma \sqrt{d}}x^{(k)}\right)}}\\
    \leq & \prod_{k=L}^{K-M-1}\exp\left(-\frac{B^{(k)}}{\gamma \sqrt{d}}x^{(k)}\right)\abs{\frac{\prod_{k=L}^{K-M-1} \round{\theta^{(k)}\exp\left(-\frac{B^{(k)}}{\gamma \sqrt{d}}x^{(k)}\right)}}{\prod_{k=L}^{K-M-1}\theta^{(k)}\exp\left(-\frac{B^{(k)}}{\gamma \sqrt{d}}x^{(k)}\right)} - 1}\\
    \leq & \exp\left(\frac{z + B_L}{\gamma \sqrt{d}}\right)\abs{\frac{\prod_{k=L}^{K-M-1} \round{\theta^{(k)}\exp\left(-\frac{B^{(k)}}{\gamma \sqrt{d}}x^{(k)}\right)}}{\prod_{k=L}^{K-M-1}\theta^{(k)}\exp\left(-\frac{B^{(k)}}{\gamma \sqrt{d}}x^{(k)}\right)} - 1}\\
    \leq & \exp\left(\frac{z + B_L}{\gamma \sqrt{d}}\right) \left(\frac{\prod_{k=L}^{K-M-1} \left(\theta^{(k)}\exp\left(-\frac{B^{(k)}}{\gamma \sqrt{d}}x^{(k)}\right) + \frac{1}{2}\right)}{\prod_{k=L}^{K-M-1}\theta^{(k)}\exp\left(-\frac{B^{(k)}}{\gamma \sqrt{d}}x^{(k)}\right)} - 1\right)\label{eq:error-individual}\\
    \leq & \exp\left(\frac{z + B_L}{\gamma\sqrt{d}}\right)\left(\prod_{k=L}^{K-M-1}\left(1 + \frac{\exp\left(\frac{B^{(k)}(b^{(k)}-1)}{\gamma\sqrt{d}}\right)}{2\theta^{(k)}}\right) - 1\right),\label{eq:error-uniform}
\end{align}
\end{small}such that with the constraint that $\prod_{k=L}^{K-M-1} \theta^{(k)} = \theta$, setting $\theta^{(k)} \gets {\theta^{(k)}}^* := \exp\left(\frac{B^{(k)}}{\gamma\sqrt{d}}(b^{(k)}-1)\right) \left(\theta\exp\left(-\frac{B_{K-M}-B_L}{\gamma\sqrt{d}}\right)\right)^\frac{1}{K-M-L}$ for each $i$ minimizes \eqref{eq:error-uniform} as \begin{equation}
     \exp\left(\frac{z + B_L}{\gamma\sqrt{d}}\right)\left(\left(1 + \frac{\exp\left(\frac{B_{K-M} - B_L}{(K-M-L)\gamma\sqrt{d}}\right)}{2\theta^{\frac{1}{K-M-L}}}\right)^{K-M-L} - 1\right).
\end{equation}

Summarizing above, the approximation error of $\exp\left(\frac{z}{\gamma\sqrt{d}}\right)$, namely $\varepsilon_{\exp}(z) = \case{C\exp(z), & -B_{K-M} <  z \leq 0; \\ \exp(z), & z \leq -B_{K-M}},$ for some asymptotically small coefficeint \[C:= \left(\exp\left(\frac{B_L}{(K-M-L)\gamma\sqrt{d}}\right) + \frac{\exp\left(\frac{B_{K-M}}{(K-M-L)\gamma\sqrt{d}}\right)}{2\theta^{\frac{1}{K-M-L}}}\right)^{K-M-L} - 1.\]

Therefore, consider a row $\mathbf{z} = (z_0, z_1, \dots, z_{n-1})$ where the Softmax is applied, due to the additional error incurred by the rounding of $\Hat{z}$ defined by \eqref{eq:shift-const}, the $L_1$-error of approximation is upper-bounded by \begin{align}
    ~ & \sum_{i=0}^{n-1} \left(\abs{\exp\left(\frac{z_i - \Hat{z}}{\gamma\sqrt{d}}\right) - \exp\left(\frac{z_i - \round{\Hat{z}}}{\gamma\sqrt{d}}\right)} + \varepsilon_{\exp}(z_i - \round{\Hat{z}})\right)\\
    \leq & \exp\left(\frac{1}{2\gamma\sqrt{d}}\right) + \sum_{i=0}^{n-1} \varepsilon_{\exp}(z_i - \round{\Hat{z}})\nonumber\\
    \leq & \left(\exp\left(\frac{1}{2\gamma\sqrt{d}}\right)-1\right) \sum_{i=0}^{n-1}\exp\left(\frac{z_i-\Hat{z}}{\gamma\sqrt{d}}\right)\nonumber \\&+ \sum_{z_i - \round{\Hat{z}}\leq -B_{K-M}}\varepsilon_{\exp}(z_i - \round{\Hat{z}}) + \sum_{z_i - \round{\Hat{z}} > -B_{K-M}}\varepsilon_{\exp}(z_i - \round{\Hat{z}}) \nonumber\\
    \leq & C \exp\left(\frac{1}{2\gamma\sqrt{d}}\right) + (n-1)\exp\left(-\frac{B_{K-M}}{\gamma\sqrt{d}}\right) =: \varepsilon_\text{attn}\label{eq:error-bound},
\end{align} such that with other variables fixed, $\varepsilon_\text{attn}$ should be minimized as a function $B_{K-M}$ which gives an optimal $B_{K-M}^*$.

Subsequently, with the choice of \begin{equation}
    B_{K-M}^*\gets \frac{\gamma\sqrt{d}}{K-M-L+1} \left(\left(K-M-L\right)\ln \left(2n\right) + \ln\theta\right),
\end{equation} the error bound in \eqref{eq:error-bound} can be minimized as \begin{equation}
    \varepsilon_\text{attn} = O\left(\left(K-M-L\right)\left(\frac{n}{\theta}\right)^\frac{1}{K-M-L+1}\right).
\end{equation}
\end{proof}

\subsection{Appendix on Security and Privacy Analysis}\label{appendix:sp-anal}

In this appendix, we present the details of the security and privacy analysis omitted in Section. \ref{sec:sp-anal}, giving the proofs of Theorems \ref{thm:tlookup-completeness} and \ref{thm:tlookup-soundness}. 

\begin{proof}[Proof of Theorem \ref{thm:tlookup-completeness}]
    If $\mathbf{S}\subset \mathbf{T}$ as sets, where $\mathbf{T} \in \mathbb{F}^N$, then in Line \ref{alg-line:tlookup-random-challenge} of Protocol \ref{protocol:tlookup}, the random challenge $\mathcal{V}$ sent to $\mathcal{P}$, namely $\beta$, is excluded from the set $E := \{x \in \mathbb{F}: -x \in \mathbf{S} \lor -\mathbf{x} \in \mathbf{T}\}$ with probability $1-\frac{N}{\abs{\mathbb{F}}}$ since $\abs{E} \leq N$. Therefore, with probability at least $1-\frac{N}{\abs{\mathbb{F}}}$, all $\beta + \mathbf{S}_i$s ($i\in \bracket{D}$) and $\mathbf{\beta} + \mathbf{T}_i$ ($i \in \bracket{N}$) are invertible, and \eqref{eq:hab22-check} holds with the definitions of $\mathbf{m}$ and $\mathbf{A}, \mathbf{B}$ as in \eqref{eq:hab22-coefs} and \eqref{eq:hab22-invs}. Therefore, by denoting the RHS of \eqref{eq:tlookup-sumcheck-preliminary} as a quadratic formula of $\alpha$, i.e., $c_0 + c_1 \alpha + c_2 \alpha^2$, we have $c_0 = 0$. Moreover, the definitions of $\mathbf{A}$ and $\mathbf{B}$ automatically guarantee that $c_1 = c_2 = 0$. Therefore, with probability $1-\frac{N}{\abs{\mathbb{F}}}$, the equality of \eqref{eq:tlookup-sumcheck-preliminary} (and therefore \eqref{eq:tlookup-sumcheck}) holds, such that the semi-honest verifier accepts the proof due to the perfect completeness of the sumcheck protocol.
\end{proof}

\begin{proof}[Proof of Theorem \ref{thm:tlookup-soundness}]
    Assume the acceptance of the proof by the semi-honest verifier \(\mathcal{V}\). Therefore:

    \begin{itemize}
        \item In Line \ref{alg-line:tlookup-sumchecks} of Protocol \ref{protocol:tlookup}, by the soundness of the Pedersen commitment scheme \cite{pedersen} (instantiated by Hyrax \cite{hyrax}), with probability \(1 - \text{negl}(\lambda)\), the prover \(\mathcal{P}\) has computed the transmitted \(\dbbracket{\mathbf{A}}\), \(\dbbracket{\mathbf{B}}\), \(\dbbracket{\mathbf{S}}\), and \(\dbbracket{\mathbf{m}}\) with correct multilinear extension values \(\me{\mathbf{A}}{\mathbf{v}}\), \(\me{\mathbf{B}}{\mathbf{v}_{\left[\log_2\frac{D}{N}:\right]}}\), \(\me{\mathbf{S}}{\mathbf{v}}\), \(\me{\mathbf{m}}{\mathbf{v}_{\left[\log_2\frac{D}{N}:\right]}}\) subject to the random challenges \(\mathbf{v}\) chosen by the verifier \(\mathcal{V}\) during the execution of the sumcheck protocol. Also, with probability \(1 - \tt{negl}(\lambda)\), the claimed value of \(\me{\mathbf{T}}{\mathbf{v}_{\left[\log_2\frac{D}{N}:\right]}}\) is correct with respect to the committed \(\mathbf{T}\) which \(\mathcal{P}\) and \(\mathcal{V}\) agree upon. 

        \item By the soundness of the sumcheck protocol, with probability \(1 - O\left(\frac{D}{\abs{\mathbb{F}}}\right)\), the correctness of \(\me{\mathbf{A}}{\mathbf{v}}\), \(\me{\mathbf{B}}{\mathbf{v}_{\left[\log_2\frac{D}{N}:\right]}}\), \(\me{\mathbf{S}}{\mathbf{v}}\), \(\me{\mathbf{m}}{\mathbf{v}_{\left[\log_2\frac{D}{N}:\right]}}\), and \(\me{\mathbf{T}}{\mathbf{v}_{\left[\log_2\frac{D}{N}:\right]}}\) implies the equality of \eqref{eq:tlookup-sumcheck}, and therefore the equality of \eqref{eq:tlookup-sumcheck-preliminary}.
        
        \item By the Schwartz-Zippel Lemma \cite{DBLP:journals/jacm/Schwartz80, DBLP:conf/eurosam/Zippel79}, the equality of \eqref{eq:tlookup-sumcheck-preliminary} implies that each term on the RHS of \eqref{eq:tlookup-sumcheck-preliminary} is \(0\) with \(1 - \frac{2}{\abs{\mathbb{F}}}\) probability. Therefore, the equality of \eqref{eq:hab22} holds for \(X \gets \beta\).
        
        \item Upon applying the Schwartz-Zippel Lemma once more, given the randomness of \(\beta\), if the equality in \eqref{eq:hab22} is valid for \(X \gets \beta\), then the equality as rational functions is also valid with a probability of \(1 - \frac{1}{DN}\). This, in turn, implies the inclusion relation \(\mathbf{S} \subset \mathbf{T}\).
    \end{itemize} 

    Combining the arguments above, under our assumption that both $\frac{D}{\abs{\mathbb{F}}}$ and $\frac{N}{\abs{\mathbb{F}}}$ are both negligible in $\lambda$, the acceptance of the proof implies $\mathbf{S} \subset \mathbf{T}$ with probability $1-\negl{\lambda}$.
\end{proof}

\end{document}